\documentclass[11pt,a4paper]{article}
\usepackage[hyperref]{acl2021}
\usepackage{times}
\usepackage{latexsym}
\usepackage{stfloats}

\renewcommand{\UrlFont}{\ttfamily\small}
\usepackage{amsmath,amsthm,amssymb}

\usepackage{microtype}
\usepackage{balance}
\usepackage{paralist}
\usepackage{verbatim}
\usepackage[normalem]{ulem}
\usepackage[vlined,ruled,linesnumbered]{algorithm2e}
\usepackage{makecell}
\usepackage{txfonts}
\usepackage{xspace}
\usepackage{booktabs}
\usepackage{graphicx}
\usepackage{xcolor,colortbl}
\usepackage{color,soul}
\usepackage{multirow}

\usepackage[T1]{fontenc}
\usepackage{bbold}
\usepackage{pifont}% http://ctan.org/pkg/pifont

\aclfinalcopy % Uncomment this line for the final submission
\def\aclpaperid{183} %  Enter the acl Paper ID here

\DeclareMathOperator*{\argmax}{arg\,max}

\DeclareMathOperator*{\dist}{D_f}
\DeclareMathOperator*{\fp}{f_p}
\DeclareMathOperator*{\E}{\mathbf{H}}

\newcommand{\colbox}[2]{\colorbox{#1}{#2}}

\newcommand{\veryshortarrow}[1][3pt]{\mathrel{%
   \hbox{\rule[\dimexpr\fontdimen22\textfont2-.2pt\relax]{#1}{.4pt}}%
   \mkern-4mu\hbox{\usefont{U}{lasy}{m}{n}\symbol{41}}}}

\definecolor{caddback}{rgb}{0.90, 0.98, 0.96}
\definecolor{cadd}{rgb}{0, 0.47, 0.34}
\definecolor{cdelback}{rgb}{1, 0.94, 0.92}
\definecolor{cdel}{rgb}{0.83, 0.32, 0.16}
\def \arrow{$\veryshortarrow$}
\newcommand{\add}[1]{\colbox{caddback}{\color{cadd}#1\xspace}} %
\newcommand{\addplus}[1]{{\footnotesize+}\add{#1}}

\newcommand{\remove}[1]{\colbox{cdelback}{{\color{cdel}#1\xspace}}}%
\newcommand{\swap}[2]{\remove{#1}~\arrow~\add{#2}}
\newcommand{\ctrltag}[1]{\texttt{#1}\xspace}

\definecolor{cexample}{rgb}{0.23, 0.30, 0.45}
\newcommand{\exinline}[1]{{\color{cexample}``#1''\xspace}}

\definecolor{ctemplate}{rgb}{0.23, 0.30, 0.45}
\definecolor{cword}{rgb}{0, 0, 0.7}
\newcommand{\BLANK}{\texttt{BLANK}}

\newcommand{\sysname}{\textsc{Polyjuice}\xspace}
\newcommand{\modelurl}{\url{https://huggingface.co/uw-hai/polyjuice}}
\renewcommand{\modelurl}{\url{https://github.com/tongshuangwu/polyjuice}}

\newcommand{\tagstr}{control code\xspace}
\newcommand{\Tagstr}{Control code\xspace}
\newcommand{\tagstrs}{control codes\xspace}
\newcommand{\tagstrshorts}{codes\xspace}
\newcommand{\tagstrshort}{code\xspace}

\newcommand{\eg}{\emph{e.g.,}\xspace}%
\newcommand{\ie}{\emph{i.e.,}\xspace}

\newcommand{\nli}{\emph{NLI}\xspace}
\newcommand{\sst}{\emph{Sentiment}\xspace}
\newcommand{\qqp}{\emph{QQP}\xspace}

\def \dsst{SST-2\xspace}
\def \dqqp{QQP\xspace}
\def \dnli{SNLI\xspace}

\newcommand{\xp}{\hat{x}}

\newcommand{\xset}{\mathbf{X}}
\newcommand{\tofix}[1]{{\color{red}#1}\xspace}
\newcommand{\fixed}[1]{{\color{blue}#1}\xspace}
\renewcommand{\fixed}[1]{{#1}\xspace}

\newcommand{\ensuretext}[1]{#1}
\newcommand{\marker}[2]{\ensuremath{^{\textsc{#1}}_{\textsc{#2}}}}
\newcommand{\arkcomment}[3]{\ensuretext{\textcolor{#3}{[#1 #2]}}}
\newcommand{\wts}[1]{\arkcomment{\marker{S}{W}}{#1}{orange}}

\title{\sysname: Generating Counterfactuals \\for Explaining, Evaluating, and Improving Models}

\author{
\makecell{
Tongshuang Wu$^{1}$ ~~~~~~~ 
Marco Tulio Ribeiro$^{2}$ ~~~~~~~ 
Jeffrey Heer$^{1}$ ~~~~~ 
Daniel S. Weld$^{1,3}$}  \\ 
$^{1}$University of Washington\hspace{5mm}
$^{2}$Microsoft Research\hspace{5mm} 
$^{3}$Allen Institute for Artificial Intelligence\hspace{5mm}\\
\hspace{-3mm}
\href{mailto:wtshuang@cs.uw.edu}{\texttt {wtshuang@cs.uw.edu}}
\hspace{4mm}
\href{mailto:marcotcr@microsoft.com}{\texttt {marcotcr@microsoft.com}}
\hspace{4mm}
\href{mailto:dan@cs.washington.edu}{\texttt {\{jheer,weld\}@cs.uw.edu}}
}

\date{}

\begin{document}
\maketitle
\begin{abstract}

While counterfactual examples
%(or perturbations) \dq{} 
are useful for analysis and training of NLP models, current generation methods either rely on manual labor to create very few counterfactuals, or 
%only instantiate a small set of automatic perturbations such as paraphrases or word substitutions.
only instantiate limited types of perturbations such as paraphrases or word substitutions. %, lacking variety.
We present \sysname, a general-purpose counterfactual generator that allows for control over perturbation types and locations, trained by finetuning GPT-2 on multiple datasets of paired sentences. 
We show that \sysname produces diverse sets of realistic counterfactuals, which in turn are useful in various distinct applications: improving training and evaluation on three different tasks (with around 70\% less annotation effort than manual generation), augmenting state-of-the-art explanation techniques, and supporting systematic counterfactual error analysis by revealing behaviors easily missed by human experts.

\end{abstract}

\section{Introduction}
\label{sec:intro}

\begin{figure}[t]
\centering
\includegraphics[trim={0 18cm 30.5cm 0cm},clip, width=1\columnwidth]{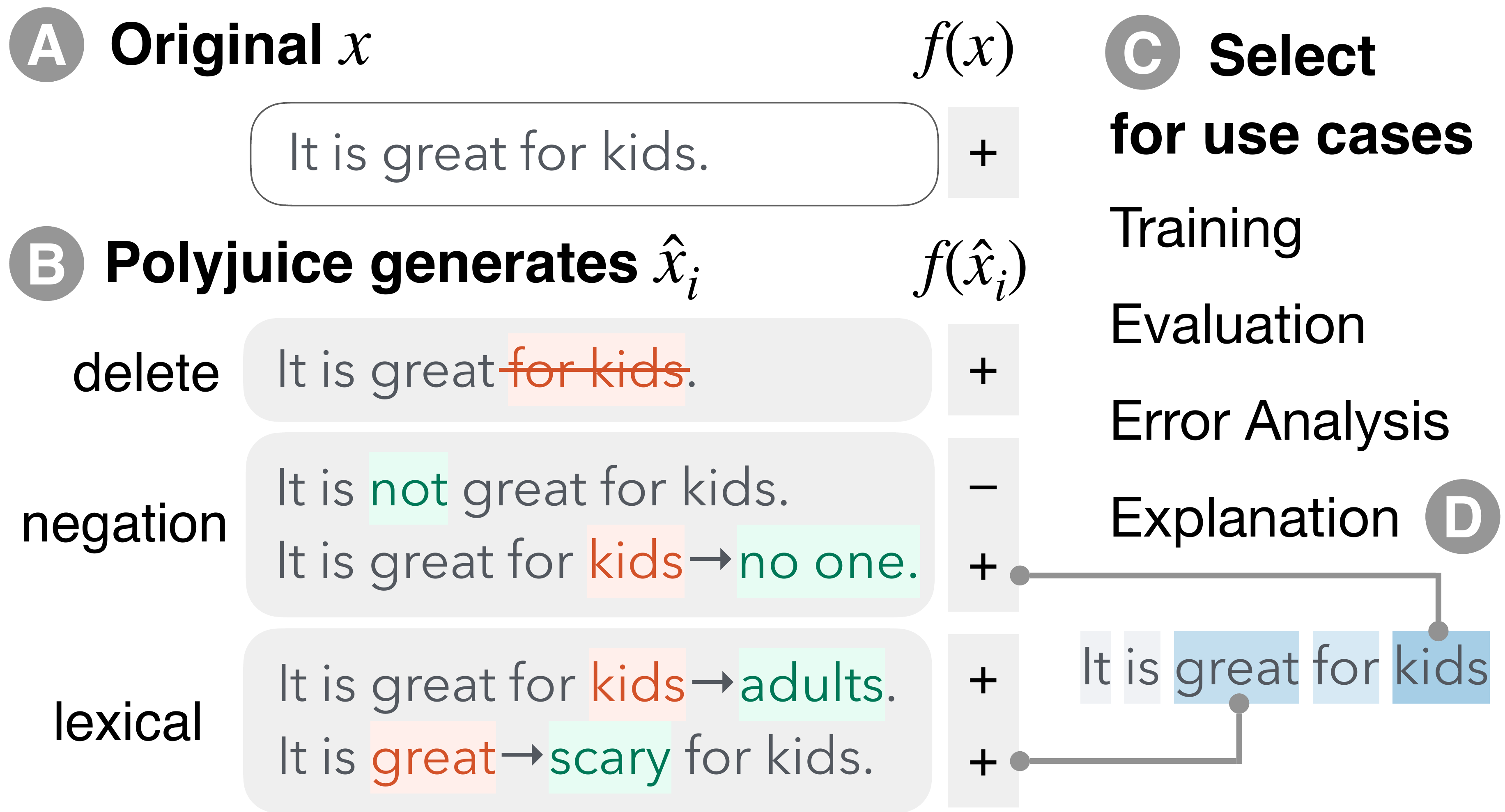}
\vspace{-5pt}
\caption{
Overview: (A) given a sentiment analysis instance $x$, \sysname\footnotemark generates (B) various counterfactuals $\xp$, which are then (C) selected for downstream use.
\eg in (D) we select counterfactual explanations that complement a black box explanation: though ``great'' and ``kids'' are deemed important, perturbing them may not affect the prediction $f(x)=f(\xp)=\text{\emph{positive}}$, revealing model failures not covered by feature attributions.
}
\vspace{-5pt}
\label{fig:teaser}
\end{figure}

Counterfactual reasoning --- mentally simulating what \emph{would have happened} if conditions were different --- is a common tool for making causality assessments~\cite{kahneman}, which in turn are crucial for model evaluation, error analysis, and explanation~\cite{miller}. 
For example, in Figure~\ref{fig:teaser}, \exinline{It is great for kids} is perturbed into multiple variations, each providing unique insights by simulating what would have happened if the sentence was different.

Applications of counterfactual reasoning to NLP generally specify the relationship $x \veryshortarrow \xp$, and then create $\xp$ according to the relationship.
As a result, prior work has tailored counterfactual generators for different applications, only collecting subsets of $\xp$ that are useful for the specific task.
%inducing different limitations.
%For example, a minimal edit of a sentence $x$ that results in a different label is useful for model training and evaluation.
%Such counterfactuals are usually produced manually by human annotators ~\cite{gardner2020contrast} or by human-written perturbation functions~\cite{wu2019errudite}, making them costly to generate (\eg 4-5 minutes per counterfactual~\cite{kaushik2019learning}) and liable to systematic omissions (\eg human annotators may miss \swap{kids}{no one} in Figure~\ref{fig:teaser}B). %Further, relying on human creativity may cause important but non-obvious patterns to be missed, \eg humans may cover \swap{great}{not great}, but miss \swap{kids}{no one} in Figure~\ref{fig:teaser}B.
%Though it is cheaper to automate the process with parsing templates~\cite{li2020linguistically}, the templates usually have limited coverage on either the patterns-to-perturb, or the applicable data points.
%In contrast, adversarial examples specify a very different counterfactual relationship: $x$ and $\xp$ must have different model predictions \emph{despite} being semantically equivalent, such that automated methods often rely on paraphrasing models~\cite{iyyer2018adversarial,  ribeiro2018semantically}.
For example, to support {\em model training and evaluation}, human annotators create counterfactuals that change the groundtruth labels by manually rewriting instances~\cite{gardner2020contrast, qin-etal-2019-counterfactual} or defining perturbation functions~\cite{checklist:acl20}.
Manual rewrites are costly (\eg 4--5 minutes per counterfactual~\cite{kaushik2019learning}) and susceptible to systematic omissions (\eg human annotators may cover \swap{great}{not great}, but miss \swap{kids}{no one} in Figure~\ref{fig:teaser}B).
Meanwhile, automated generators for \emph{model analysis and explanation} usually focus on other relationships, \eg generating $\xp$ that have different model predictions than $x$~\cite{ross2020explaining, Zhang2019GeneratingFA}. 
As a result, they neglect prediction-preserving counterfactuals that are equally important for understanding or shaping model behaviors, like \swap{kids}{no one} and \swap{great}{scary} linked to Figure~\ref{fig:teaser}D.

%In contrast, adversarial examples\dan{now you come to another use case, but 'adversarial examples' is 1) not a reason or use for Cfs; it's a characterization of theirs style, 2) it doesn't match the list of uses cases you've lined up above and (crucially) in fig 1, so 3) it doesn't reinforce Fig 1, which is the central framing of the paper. I suggest you rewrite to follow the use cases. It really matters to explain the motivation of thge paper simply and clearly and to repeat it (in a non annoying way) - avoids confusion} specify a different relationship: $x$ and $\xp$ must have the same groundtruth but different model predictions, driving the generator to prioritize paraphrasing~\cite{iyyer2018adversarial, ribeiro2018semantically}, over the non-paraphrasing changes that still preserve the groundtruth~\cite{li2020contextualized}.
%(\eg changing movie names in sentiment analysis).

%\footnotetext{We open source both the \sysname model and the selection heuristics at \modelurl.}
\footnotetext{We open source \sysname at \modelurl.}

%However, as Figure~\ref{fig:teaser} illustrates, counterfactual generation does not \emph{have} to be task-specific, as various applications share similar requirements on $x\veryshortarrow\xp$ (\eg a preference for small changes).
%In fact, a general-purpose pool of diverse counterfactuals may be preferable when the relationship is not precisely defined in advance, as is the case for counterfactual training, evaluation, or explanations.
% In fact, a common pool of counterfactuals may support each application more comprehensively.
% For example, without the preset constrain on semantic equivalence, we can collect adversarials through negation (for tasks insensitive to negation, \eg named entity recognition.)
%For example, some forms in training and evaluation data collection can 
%The exhaustive search of automated methods can cover the training and evaluation more comprehensively, and non-paraphrasing changes like \emph{add negation} are valuable adversarials for tasks like named entity recognition. 
%\hao{might be good to clarify that these two are not mutually-exclusive: adversarial examples should sometimes be close to the original}
% 
However, counterfactual generation does not \emph{have} to be task-specific.
The same set of counterfactuals in Figure~\ref{fig:teaser} can support a variety of applications.
Moreover, for cases like model explanation and analysis, a general-purpose pool of counterfactuals may be preferable, as the relationship of interest can be more exploratory and user-oriented~\cite{wu2019errudite}.
In this work, we formalize the task of \emph{counterfactual generation}, disentangling generation from the application of counterfactuals.
Given an input $x$ (Figure~\ref{fig:teaser}A), our generator produces a set of counterfactuals $\hat{\xset} = \{\xp_1, \xp_2, ...\}$ with \emph{application-agnostic} relationships $x \veryshortarrow \xp_i$ (Figure~\ref{fig:teaser}B).
Afterwards, we use \emph{application-specific} selection methods to find subsets of $\xp$ that are most effective for a given use case (Figure~\ref{fig:teaser}C).

We frame the generation step as conditional text generation, and finetune GPT-2~\cite{radford2019language} into a generator called \emph{\sysname} using $(x, \xp)$ pairs. 
To allow for targeted counterfactuals, we also design \tagstrs like \ctrltag{negation} or \ctrltag{delete} (Figure~\ref{fig:teaser}B), and adopt fill-in-the-blank structures~\cite{donahue2020enabling} to specify where the perturbation occurs and how.
%We also allow for targeted counterfactuals, by specifying where the perturbation occurs~\cite{donahue2020enabling} and designing \tagstrs like \ctrltag{negation} or \ctrltag{delete} (Figure~\ref{fig:teaser}B).
Intrinsic evaluation shows that \sysname generates $\xp$ that are \emph{fluent}, \emph{diverse}, and \emph{close to $x$}, and that the \emph{control} mechanisms retrieve perturbations that would likely not be sampled from off-the-shelf language models. %(\eg 42\% more negations).

%--- the control is \emph{the backbone of} various downstream applications.

%We propose simple yet effective selection strategies, 
With simple selection heuristics, we show that a single \sysname model can significantly aid humans in diverse downstream applications.\footnote{We demonstrate \sysname in semi-automatic settings, but as discussed in \S\ref{subsec:nlg}, it can also work automatically.} 
For \emph{counterfactual training and evaluation} (\S\ref{sec:app_label}), humans label \sysname counterfactuals rather than creating them from scratch.
They produce training data that significantly improve model generalization, as well as contrast sets that help identify model vulnerabilities~\cite{gardner2020contrast}, with around 70\% less annotation effort. 
In another application, \sysname produces \emph{counterfactual explanations} (\S\ref{sec:app_explain}), providing significant insight on top of state-of-the-art explanation techniques. 
Finally, \sysname supports counterfactual \emph{error analysis} (\S\ref{sec:app_err_analysis}).
It allows users to explore related counterfactuals (\eg the model responds differently to different negation forms in Figure~\ref{fig:teaser}B), and to aggregate individual counterfactuals into patterns in order to gain systematic understanding of model behavior.

\section{General-Purpose Counterfactuals}
\label{sec:general_purpose}

\newcommand{\tagdefine}[1]{\emph{{\color{darkgray}#1} }}
\newcommand{\TagTable}{
\begin{table*}
\small
\centering
\begin{tabular}{@{} p{0.11\linewidth} p{0.61\linewidth} p{0.22\linewidth} @{}}
\toprule
\textbf{\Tagstr} & \textbf{Definitions and \sysname-generated Examples} & \textbf{Training Datasets} \\ 
\midrule
\ctrltag{negation}
 & A dog is \add{not} embraced by the woman.
 &\cite{kaushik2019learning}
\\ \midrule
\ctrltag{quantifier}
 & \swap{A dog is}{Three dogs are} embraced by the woman. 
 &\cite{gardner2020contrast}
\\ \midrule
\ctrltag{shuffle}
 & \tagdefine{To move (or swap) key phrases or entities around the sentence.} \newline
 A \swap{dog}{woman} is embraced by the \swap{woman}{dog}.
 &\cite{zhang2019paws}
\\ \midrule
\ctrltag{lexical}
 & \tagdefine{To change just one word or noun chunk without altering the POS tags.} \newline
 A dog is \swap{embraced}{attacked} by the woman.
 &\cite{sakaguchi2019winogrande}
\\ \midrule
\ctrltag{resemantic}
 & \tagdefine{To replace short phrases without altering the remaining dependency tree.}\newline
 A dog is \swap{embraced by the woman}{wrapped in a blanket}.
 &\cite{wieting2017paranmt}
\\ \midrule
\ctrltag{insert}
 & \tagdefine{To add short phrases without altering the remaining dependency tree.} \newline
 A dog is embraced by the \add{little} woman.
 &\cite{mccoy2019right}
\\ \midrule
\ctrltag{delete}
 & \tagdefine{To remove short phrases without altering the remaining dependency tree.} \newline
 A dog is embraced \remove{by the woman}.
 &\cite{mccoy2019right}
\\ \midrule
\ctrltag{restructure}
 & \tagdefine{To alter the dependency tree structure, \eg changing from passive to active.} \newline
 A dog is \swap{embraced by}{hugging} the woman.
 &\cite{wieting2017paranmt}
\\
\bottomrule
\end{tabular}
\vspace{-5pt}
\caption{
We design a list of \tagstrs to guide generation.
We show \emph{\sysname-generated} counterfactual examples, and the representative training datasets for each corresponding pattern. 
Details are in Appendix~\ref{appendix:train_data}.
%More examples are in Appendix~\ref{appendix:example}.
}
%\wts{Change all the examples to be on an identical sentence, not all different cases. And consider further annotate the tags based on whether they just do semantic change or also syntactic change.}}
\label{table:ctrltag}
%\vspace{-10pt}
\end{table*}
}
% on a particular instance
% 
\subsection{Definition and Desiderata}
\label{sec:desiderata}

Given an instance $x$, a generator $g$ produces a set of counterfactuals $\hat{\xset} = \{\xp_1, \xp_2, ...\}$ with various relationships $x \veryshortarrow \xp_i$. % (referred as $\relation{\xp_i}$ for simplicity).
% Each $\xp_i$ perturbs $x$ with certain strategies like negations, syntactic restructuring, etc., and the edited spans are instantiations of the strategies.
For example, \swap{great}{not great}, \swap{kids}{no one} in Figure~\ref{fig:teaser}B are both instances of the \ctrltag{negation} relationship.
% The importance of certain $\relation{\xp}$ changes along with applications.
Each $(x, \xp)$ pair shares multiple relationships --- these two are also instances of the \emph{label flipping} relationship if the task is sentiment analysis (but might not be for other tasks).
As illustrated in \S\ref{sec:intro}, knowing which relationships apply aids selection for downstream applications.

We expect $g$ to produce counterfactuals $\xp$ that are (1) \textbf{close} to $x$, preferably only involving the minimal changes necessary to establish a certain effect~\cite{pearl2018causal}, allowing users to make causality assessments.
The generated $\xp$ should also be (2) \textbf{fluent}, \ie grammatically correct~\cite{morris2020textattack} and semantically meaningful (\eg \exinline{Colorless green ideas sleep furiously} is not meaningful~\cite{chomsky2002syntactic}).
Fluency operationalizes ``probable'' counterfactuals in the context of NLP;
as \citet{kahneman} stated, humans strongly favor counterfactuals that are close to the original instance, but also prefer those that could have easily happened without assuming rare events or strange coincidences.
%Humans strongly favor counterfactuals that are close to the original instance, but also prefer them to be probable, \ie they could have easily happened without assuming rare events or strange coincidences~\cite{kahneman}.
%We operationalize ``probable'' in the context of NLP by requiring $g$ to generate (2) \textbf{fluent} $\xp$, \ie grammatically correct~\cite{morris2020textattack} and semantically meaningful (\eg \exinline{Colorless green ideas sleep furiously} is not meaningful~\cite{chomsky2002syntactic}).
Further, as a general-purpose generator, $g$ should produce counterfactuals with a measure of (3) \textbf{control} over relationships $x \veryshortarrow \xp$,  %$\relation{\xp}$,
such that the counterfactuals can vary with the object-of-attention in each application (the ``focus rule''~\cite{kahneman}).
Finally, %even though infinite diverse counterfactuals can be produced~\cite{pearl2018causal},
we expect $g$ to output a (4) \textbf{diverse} set of $\xp$ in terms of relationships, covering a large variety of ``what-ifs'' for different applications~\cite{pearl2018causal}.

\TagTable

\begin{figure}[t]
\centering
\includegraphics[trim={0 18.6cm 31.5cm 0cm}, clip, width=1\columnwidth]{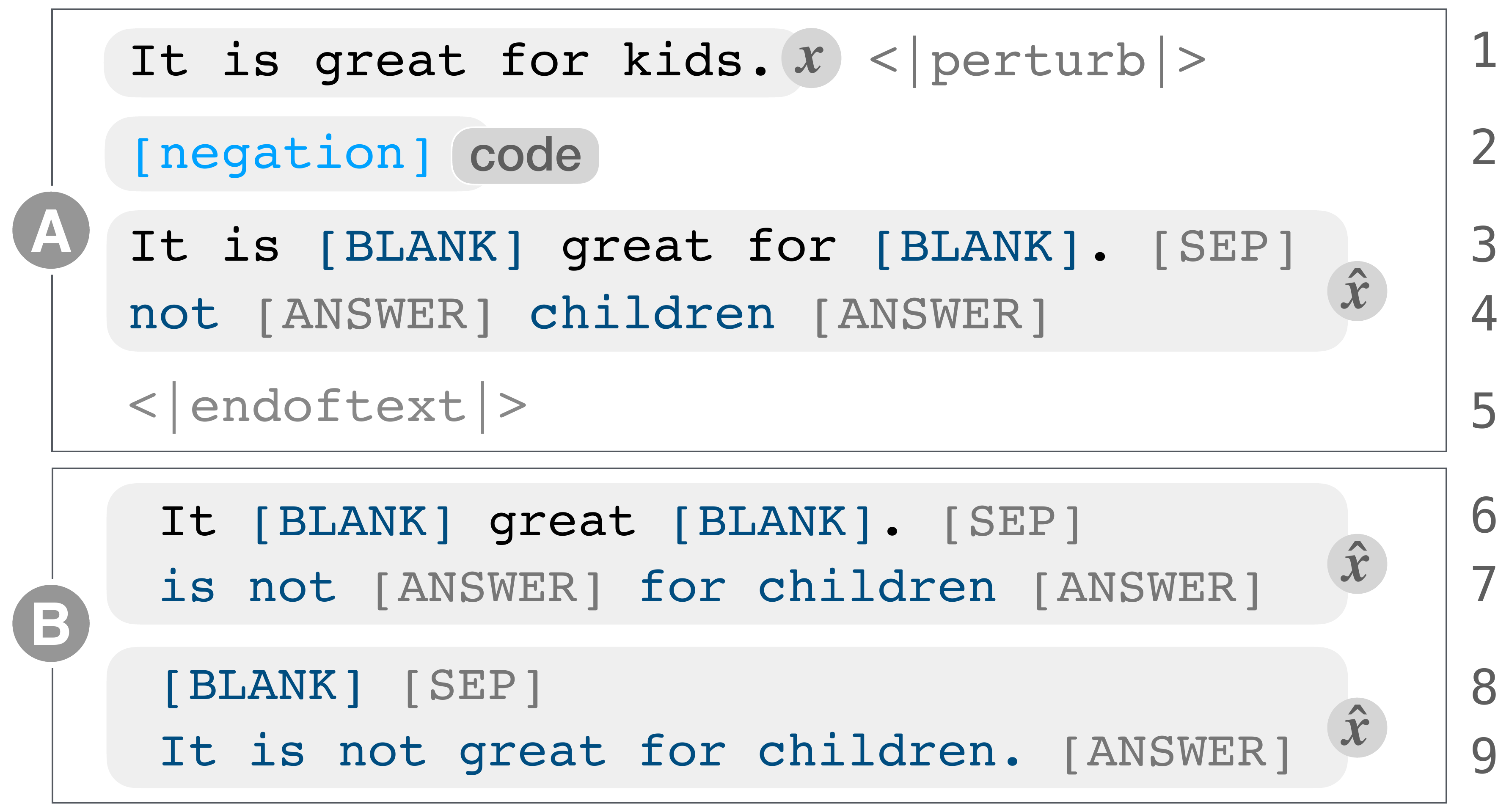}
\vspace{-15pt}
\caption{ 
(A) \sysname prompt format, which concatenates the original $x$, the \tagstr, and the $\xp$ (\exinline{It is not great for children} converted to an infilling structure).
At \emph{generation} time, \sysname accepts prompts that just include $x$ (Line 1), or optionally with the \tagstrshort and the \texttt{[BLANK]}s (Lines 2--3), and fills in the blanks sequentially with spans separated by \texttt{[ANSWER]}s (Line 4).
%During training, the \tagstrshorts and blanks are automatically extracted.
(B) \sysname allows blanking at different granularities (even the entire sentence), such that Lines 3--4 in (A) can be replaced by Lines 6--7 or 8--9. 
%\dq{}
%We get multiple training texts per pair, by blanking $\xp$ subtrees that contain the change, or the entire sentence.
}
\vspace{-10pt}
\label{fig:blank}
\end{figure}

\subsection{Conditional Counterfactual Generation}
\label{subsec:nlg}

%We frame counterfactual generation as a conditional text generation task using language models (LMs).
We frame counterfactual generation as a conditional text generation task using language models (LMs)\fixed{, and train \sysname by finetuning GPT-2~\cite{radford2019language} using the following prompt design (alternative LMs could also have been used).}

\paragraph{\fixed{Prompt format design.}}
To ensure that $\xp$ is \emph{close} to $x$ rather than arbitrary text, we condition the generation on $x$, followed by a special token (Line 1 in Figure~\ref{fig:blank}A).
In Line 2, we have \emph{\tagstrs} \cite{ctrl} such as \ctrltag{negation}.
We design them to specify types of perturbation from among lexical, syntactic, or semantic aspects (see Table \ref{table:ctrltag}), inspired by prior work that categorizes manually created counterfactuals~\cite{kaushik2019learning, gardner2020contrast}.
% All the \tagstrshorts are described in Table~\ref{table:ctrltag}. 
As an additional layer of control over $x \veryshortarrow \xp$, % $\relation{\xp}$,
we allow users to specify \emph{where} changes happen by having the LM infill \texttt{[BLANK]} tokens~\cite{donahue2020enabling}, rather than generating arbitrary counterfactuals (Lines 3--4).
%At generation time, if the user provides only the original example, \sysname will generate the \tagstr, the blank locations, and the infilling (Lines 2--4). 
%Alternatively, the user can specify the \tagstr, or the \tagstr \emph{and} the location of the blanks, to exercise different degrees of control depending on the application.

\fixed{Finetuning GPT-2 --- a causal LM for predicting next tokens --- additionally allows us to exercise control at various levels of granularity.
At generation time, if the user provides only the original example, \sysname will generate the \tagstr, the blank locations, and the infilling (Lines 2--4).
Alternatively, the user can specify the \tagstr, or the \tagstr \emph{and} the blanks, to exercise different degrees of control depending on the application.
As later shown in \S\ref{sec:app_explain} and \S\ref{sec:app_err_analysis}, such control is important for different use cases.}

\paragraph{\fixed{Training data.}}
To train a conditional model, we combine six existing sentence-pair datasets, each containing a subset of the desired phenomena in Table~\ref{table:ctrltag}. 
Further, we find naturally occurring sentence pairs (filtered by edit distance to guarantee closeness) in non-paired datasets including CommonGen~\cite{lin-etal-2020-commongen}, Natural Questions~\cite{kwiatkowski-etal-2019-natural}, and SQuAD~\cite{rajpurkar-etal-2016-squad}, such that the resulting dataset contains \emph{diverse} counterfactuals.\footnote{We exclude data related to our applications, \eg PAWS-QQP \cite{zhang2019paws}. }
%(details in Appendix~\ref{appendix:train_data}). 
%We translate each $(x, \xp)$ in the dataset into the format given in Figure~\ref{fig:blank}A, by computing the \tagstr from syntactic features (POS tags and dependency trees), and placing \texttt{[BLANK]}s in $\xp$. 
%To allow flexible blanking at generation time, we create multiple prompts per pair that cover different dependency tree structures related to the perturbed spans (Figure~\ref{fig:blank}B), resulting in $657,144$ prompts from $186,451$ pairs.

\fixed{We translate these sentence pairs into the format given in Figure~\ref{fig:blank}A.
For each $(x, \xp)$, we compute its primary \tagstr using part-of-speech tags and dependency trees.
%, and blank out the changed subtrees in $\xp$.
For example, \ctrltag{negation} occurs when we observe changes to negation modifiers or specific words like ``supposedly'', and \ctrltag{shuffle} occurs when we have overlap between tokens deleted and added.
When multiple changes occur, we label it with the \tagstr which most significantly changes the semantics of the corresponding subphrase as computed by SBERT~\cite{reimers-2019-sentence-bert}.
For example, in Figure~\ref{fig:blank}A, \ctrltag{negation} (\swap{great}{not great}) is more significant than \ctrltag{lexical} (\swap{kids}{children}).
To balance the distribution (Table~\ref{table:gpt_train_stats} in Appendix~\ref{appendix:train_data}), for each dataset, we extract \tagstrs from all the $(x, \xp)$,\footnote{We use sentences in a pair interchangeably as $x$ and $\xp$ to learn the \tagstrs both ways.} and randomly sample up to 10,000 instances per \tagstrshorts.
%Still, \ctrltag{quantifier} and \ctrltag{negation} have less training data compared to other codes. 
%Fortunately, these codes tend to be limited to more specific patterns (``more than'', ``not'', ``never'') when compared to ``broad'' codes like \ctrltag{lexical}, and thus even a small sample is enough to learn them. 

In order to allow for flexible blanking at generation time, we generate multiple training prompts per pair, covering different dependency tree structures related to the perturbed spans (Figure~\ref{fig:blank}B), including (1) just the changed tokens, (2) the associated parsing structures, (3) the merged changes, and (4) the entire sentence.
We eventually obtain $657,144$ prompts from $186,451$ pairs.}

\paragraph{\fixed{Fluency filtering.}}
While the original GPT-2 produces \emph{fluent} text, some combinations of \tagstrs and blanks cause \sysname to generate nonsensical results.
Following \citet{morris2020textattack}, we score both $x$ and $\xp$ with GPT-2, and filter $\xp$ when the log-probability (on the full sentence or the perturbed chunks) decreases by more than 10 points relative to $x$.
Fully automated uses of \sysname (\eg adversarial attacks) may benefit from stricter constraints, at the cost of diversity (as surprising changes may be filtered even if they are fluent).

\subsection{Intrinsic Evaluation}
\label{subsec:intrinsic}

\begin{table}[tb]
\small
    \centering
    \begin{tabular}{@{}lccc@{}}
    \toprule
    \multirow{2}{*}{Model} & Diversity & \multicolumn{2}{c}{Closeness} \\
    \cmidrule(lr){2-2}
    \cmidrule(lr){3-4}
    & Self-BLEU $\downarrow$ & Levenshtein $\downarrow$ & Syntactic $\downarrow$ \\
    % \cmidrule{2-4}
    \midrule
    \emph{\sysname} & 0.34     & \textbf{0.25} & \textbf{2.13} \\
    GPT-2           & \textbf{0.18}     & 0.70          & 6.35 \\
    T5              & \textbf{0.12}             & 9,52          & 3.50 \\
    RoBERTa         & 0.47              & \textbf{0.14} & \textbf{1.32} \\
    \bottomrule
    \end{tabular}
    \vspace{-5pt}
    \caption{\fixed{Intrinsic evaluations: \sysname counterfactuals are \emph{closer} to the original instance than non-fintuned GPT-2 and T5, and more \emph{diverse} than RoBERTa. Computational details are in Appendix~\ref{appendix:intrinsic}.}}
    \vspace{-5pt}
    \label{table:intrinsic}
\end{table}

\fixed{
We evaluate \sysname on \emph{closeness} and \emph{diversity} by comparing its perturbations on 300 randomly selected sentences with baselines that use more or less context from $x$: 
(1) non-finetuned GPT-2, (2) token-infilling RoBERTa~\cite{liu2019roberta} and (3) span-infilling T5~\cite{JMLR:v21:20-074}.

As shown in Table~\ref{table:intrinsic}, \sysname generates counterfactuals that are close to the original instance, measured by syntactic tree~\cite{zhang1989simple} and Levenshtein edit distance~\cite{levenshtein1966binary}.
In contrast, non-finetuned GPT-2 generates arbitrary text instead of perturbations when given the starting tokens of a sentence, as it only leverages context in a single direction. 
As for infilling models, \sysname counterfactuals are more diverse (measured by self-BLEU~\cite{zhu2018texygen}) than RoBERTa ones, which is restricted to word substitution.
Meanwhile, T5 displays higher diversity but less closeness, probably due to the fact that it does not consider the original masked tokens when generating $\xp$.
For example, in Figure~\ref{fig:teaser} \exinline{It is great for kids,} T5 replaces \exinline{for kids} with \exinline{idea}, \exinline{to meet you,} whereas \sysname generates \exinline{for kids yet adults can enjoy,} \exinline{for any audience.} 

We evaluate \emph{controllability} by comparing \sysname with T5 as well as with GPT-2 finetuned on prompts \emph{without} \tagstrshorts.
We verify that the \tagstrshorts improve the success rate of generating counterfactuals with the desired perturbation types set out in Table~\ref{table:ctrltag} by as much as 42\% for perturbations such as \ctrltag{negation} and \ctrltag{insert}.
For example, given \exinline{It is \texttt{[BLANK]} great for kids,} baselines generate \exinline{also,} \exinline{fun and,} rather than \exinline{not} (\ctrltag{negation}).

We further verify the \emph{fluency} for \sysname counterfactuals in three tasks/datasets: (1) \sst Analysis, \dsst~\cite{socher2013recursive},
(2) Natural Language Inference (\nli), \dnli~\cite{bowman-etal-2015-large}, and 
(3) Duplicate Question Detection (\dqqp)~\cite{wang2018glue}.
We randomly select 100 sentences per dataset, generate 3 $\xp$ per $x$, and ask crowd workers to rate whether they are \emph{``likely written by native speakers.''}
The workers rated most counterfactuals as fluent: $78\%$ in \dsst, $76\%$ in \dqqp, and $86\%$ in \dnli. In subsequent sections, we show these rates are suitable for applications where people ``team up'' with \sysname.
}

\newcommand{\maug}{\texttt{m-polyjuice}\xspace}
\newcommand{\mcomp}{\texttt{m-baseline}\xspace}
\newcommand{\mcad}{\texttt{m-CAD}\xspace}

\definecolor{ccon}{HTML}{fee9d4}
\definecolor{cood}{HTML}{d8f0d3}
\definecolor{cid}{HTML}{dae8f5}

\newcommand{\TableContrastSet}{
\begin{table}
\small
\centering
\setlength{\tabcolsep}{3pt}
\begin{tabular}{@{} c c c l c @{}}
\toprule
\textbf{Task} & \textbf{Dev.} & \textbf{Orig. set} & \textbf{Contrast set} $\downarrow$ & \textbf{Consistency} $\downarrow$ \\ 
\midrule
\sst & 94.3 & 93.8 & 84.9 (-8.9) & 76.1 \\
\nli & 86.5 & 91.6 & 72.3 (-19.3) & 56.4 \\
\qqp & 91.7 & 87.5 & 75.3 (-12.2) & 61.1\\
% \sst-distilbert & 91.1 & 93.2 & 84.0 (-9.2) & 75.0 \\
% \sst-bert & 92.4 & 90.9 & 85.8 (-9.2) & 76.1 \\
% \sst-roberta & 94.3 & 93.8 & 84.9 (-8.9) & 76.1 \\

% \nli-bert & 78.6 & 83.1 & 69.7 (-13.4) & 49.5 \\
% \nli-roberta & 86.5 & 91.6 & 72.3 (-19.3) & 56.4 \\
% \qqp-bert & 90.9 & 88.1 & 75.7 (-12.4) & 60.5 \\
% \qqp-distilbert & 89.7 & 88.1 & 72.4 & 57.3\\
% \qqp-roberta & 91.7 & 87.5 & 75.3 & 61.1\\
% the baseline
% \nli & 86.5 & 80.6 & 78.6 (-19.3) & 30.4 \\
% using a imdb model
% imdb_contrast_test 96.7 / 89.1 / 86.1
% imdb_iclr_test 96.7 / 91.0 / 87.9
% using a sst-2 model
% imdb_iclr_test 91.3 / 89.5 / 81.4
% imdb_iclr_test 89.5 / 87.3 / 77.3
\bottomrule
\end{tabular}
\vspace{-5pt}
\caption{\sysname $\xp$ as contrasts sets, with model accuracy on the development set, the original set of $x$, the contrast sets, and consistency (cases where the model predicts both $x$ and $\xp$ correctly).
The performance drops are similar to that of expert-created sets~\cite{gardner2020contrast}, on which the accuracy of all classification models decreases by $9.8$ on average, with a consistency of ${\approx}64.1$.
This indicates \sysname can be used to create such sets without expert annotators and at less cost.
%\ie with similar drop in performance (on average $-9.8$ for all classification tasks in \cite{gardner2020contrast}) and consistency (${\approx}64.1$ in manual contrast set).
%The model performance here drops for a similar degree compared to the manually created contrast sets~\cite{gardner2020contrast} ($-9.78$ for all )
}
\vspace{-5pt}
\label{table:contrast_set_result}
\end{table}
}
%\end{comment}

\newcommand{\TableAugSST}{
\begin{table*}
\small
\centering
\setlength{\tabcolsep}{5pt}
\begin{tabular}{@{}lllllllll@{}}
\toprule
 Model & 
 \cellcolor{cid}SST-2 & 
 \cellcolor{cood}Senti140 & 
 \cellcolor{cood}SemEval & 
 \cellcolor{cood}Amzbook & 
 \cellcolor{cood}Yelp & 
 \cellcolor{cood}IMDB & 
 \cellcolor{ccon}IMDB-Cont. & 
 \cellcolor{ccon}{IMDB-CAD} \\
\midrule
\mcomp & $92.9\pm 0.2$ & $88.9\pm 0.3$ & $84.8\pm 0.5$ & $85.1\pm 0.4$ & $90.0\pm 0.3$ & $90.8\pm 0.5$ & $92.2\pm 0.6$ & $86.5\pm 0.2$ \\
\maug	 & $92.7\pm 0.2$ & $\mathbf{90.7\pm 0.4}$ & $\mathbf{86.4\pm 0.1}$ & $85.6\pm 0.8$ & $90.1\pm 0.0$ & $90.6\pm 0.3$ & $\mathbf{94.0\pm 0.3}$ & $\mathbf{89.7\pm 0.5}$ \\
 % imdb_contrast_test: 91.1 (9.4) / 92.8 (0.4)
 % imdb_contrast_test: 87.4 (0.0) / 89.6 (0.5)
 % imdb_iclr_test 93.0 (0.3) / 93.9 (0.4)
 % imdb_iclr_dev 92.0 (0.2) / 92.7 (0.2)
\bottomrule
\end{tabular}
\vspace{-5pt}
\caption{\sst model performance, with $n{=}4,000$ and $m{=}2,000$.
\textbf{Bolded} cells highlight significant improvements.
\maug maintains the \colbox{cid}{in-domain} and \colbox{cood}{out-of-domain} accuracies on reviews (SST-2, Amzbook, Yelp, IMDb Movie Review~\cite{ni2019justifying, asghar2016yelp, maas2011learning}), improving it on Twitter data (Senti140 and SemEval 2017~\cite{go2009twitter, rosenthal2017semeval}) and \colbox{ccon}{contrast sets}~\cite{gardner2020contrast, kaushik2019learning}, likely because their distributions are less similar to the original SST-2 training data.% than the reviews.
% The model also improves on the \colbox{ccon}{contrast sets}~\cite{gardner2020contrast, kaushik2019learning}).
% we need to say it's cheaper
%on \colbox{cid}{in domain}, \colbox{cood}{out of domain}, and \colbox{ccon}{contrast sets}. \maug performs better than \mcomp on twitter datasets (Senti140~\cite{go2009twitter}, SemEval 2017~\cite{rosenthal2017semeval}) and contrast sets IMDb-Contrast~\cite{gardner2020contrast} and IMDb-CAD~\cite{kaushik2019learning}, while maintaining the ones on reviews (SST-2, Amzbook~\cite{ni2019justifying}, Yelp~\cite{asghar2016yelp}, IMDb Movie Review~\cite{maas2011learning}).
}
%\vspace{-5pt}
\label{table:aug_sst}
\end{table*}}

%%%%%%%%%%%%%%%%%%%%%%%%%%%%%%%%%%%%%%%%%%%%%%%
\newcommand{\TableAugNLI}{
\begin{table*}
\small
\centering
\setlength{\tabcolsep}{5.7pt}
\begin{tabular}{@{}lllllllll@{}}
\toprule
 Model & 
 \cellcolor{cid}SNLI & 
 \cellcolor{cood}MNLI-m & 
 \cellcolor{cood}MNLI-mm & 
 \cellcolor{ccon}SNLI-CAD & 
 \cellcolor{ccon}break & 
 \cellcolor{ccon}DNC & 
 \cellcolor{ccon}stress & 
 \cellcolor{ccon}diagnostic \\
\midrule
\mcomp 	& $85.7\pm 0.4$& $86.1\pm 0.2$& $\mathbf{86.6\pm 0.2}$& $72.8\pm 0.3$& $86.4\pm 1.5$& $54.5\pm 0.6$& $65.1\pm 0.6$& $56.0\pm 0.8$\\
 
\mcad & $85.8\pm 0.6$ & $\mathbf{86.6\pm 0.1}$ & $85.6\pm 0.3$ & $\mathbf{73.8\pm 0.2}$ & $89.4\pm 2.9$ & $55.8\pm 0.9$ & $65.5\pm 0.5$ & $56.4\pm 0.4$ \\
 
\maug	& $85.3\pm 0.3$& $86.0\pm 0.1$& $\mathbf{86.4\pm 0.0}$& $\mathbf{73.6\pm 0.2}$& $\mathbf{89.1\pm 1.2}$& $\mathbf{57.7\pm 0.3}$& $65.1\pm 0.2$& $\mathbf{57.5\pm 0.5}$\\
%  \texttt{m-polyjuice-rand} & $85.7\pm 0.4$ & $86.1\pm 0.1$ & $86.2\pm 0.1$ & $73.4\pm 0.5$ & $87.2\pm 0.6$ & $54.7\pm 0.3$ & $64.6\pm 0.6$ & $56.9\pm 0.8$ \\

 %10000 & 1574 & comp & $85.3\pm 0.5$& $85.2\pm 0.2$& $85.4\pm 0.3$& $72.4\pm 0.1$& $86.1\pm 1.8$& $54.2\pm 1.8$& $64.0\pm 0.4$& $56.0\pm 0.3$\\
 %10000 & 1574 & aug\_gpt & $85.3\pm 0.3$& $85.0\pm 0.2$& $85.1\pm 0.1$& $73.4\pm 0.5$& $90.5\pm 1.1$& $56.5\pm 1.2$& $64.6\pm 0.5$& $57.0\pm 0.4$\\
\bottomrule
\end{tabular}
\vspace{-5pt}
\caption{\nli models, with $n{=}20,000$ and $m{=}1,574$. 
\maug improves accuracy on \colbox{ccon}{contrast} and \colbox{ccon}{challenge sets}~\cite{kim2019probing, naik2018stress, glockner-etal-2018-breaking, wang2018glue}; it exhibits comparable (or better) gains than \mcad (manual counterfactuals) with less implementation and annotation effort.
}
\vspace{-5pt}
\label{table:aug_nli}
\end{table*}
}

%%%%%%%%%%%%%%%%%%%%%%%%%%%%%%%%%%%%%%%%%%%%%%%
\newcommand{\TableAugQQP}{
\begin{table}
\small
\centering
\begin{tabular}{lll}
\toprule
 Model & \cellcolor{cid}QQP & \cellcolor{ccon}PAWS-QQP \\
\midrule
\mcomp 	& $84.5 \pm 0.6$ & $37.0\pm 0.5$\\
 %20,000 & 1,574 & \texttt{aug-r} & $85.7\pm 0.4$ & $86.1\pm 0.1$ & $86.2\pm 0.1$ & $73.4\pm 0.5$ & $87.2\pm 0.6$ & $54.7\pm 0.3$ & $64.6\pm 0.6$ & $56.9\pm 0.8$ \\
\maug	& $84.7 \pm 1.0$ & $\mathbf{38.7\pm 0.4}$\\
 
 %10000 & 1574 & comp & $85.3\pm 0.5$& $85.2\pm 0.2$& $85.4\pm 0.3$& $72.4\pm 0.1$& $86.1\pm 1.8$& $54.2\pm 1.8$& $64.0\pm 0.4$& $56.0\pm 0.3$\\
 %10000 & 1574 & aug\_gpt & $85.3\pm 0.3$& $85.0\pm 0.2$& $85.1\pm 0.1$& $73.4\pm 0.5$& $90.5\pm 1.1$& $56.5\pm 1.2$& $64.6\pm 0.5$& $57.0\pm 0.4$\\
\bottomrule
\end{tabular}
\vspace{-5pt}
\caption{\sysname with $n{=}20,000$ and $m{=}1,911$ improves accuracy on \colbox{ccon}{PAWS-QQP}~\cite{zhang2019paws}.}
\vspace{-10pt}
\label{table:aug_qqp}
\end{table}
}

%%%%%%%%%%%%%%%%%%%%%%%%%%%%%%%%%%%%%%%%%%%%%%%
\newcommand{\TableAugQQPChecklist}{
\begin{table}
\small
\centering
\setlength{\tabcolsep}{4pt}
\begin{tabular}{@{}p{0.4\textwidth} r@{}}
\toprule
TESTNAME & $\Delta$ fail\% \\
\midrule
 Order does not matter for symmetric relations & -18.4\% \\
 Order does not matter for comparison & -26.5\% \\
 Order does matter for asymmetric relations & -14.5\% \\
%\midrule
% Is it \{ok, bad,..\} to \{smoke, do,..\} \{\emph{before $\not\eq$ after}\} & -52.5\% \\
% %What was person's life \{\emph{before $\not\eq$ after}\} becoming X & -46.6\% \\
% Do you have to X your dog \{\emph{before $\not\eq$ after}\} Y it & -35.4\% \\
%\midrule
% Is person X $\not\eq$ Is person becoming X & -8.5\% \\
% Is person X $\not\eq$ Did person use to be X & -5.4\% \\
\midrule
 How can I become \{\emph{more X $\not\eq$ less X}\} & -30.7\% \\
 How can I become \{\emph{more X $=$ less antonym(X)}\} & 28.0\% \\
 How can I become \{\emph{X $\not\eq$ not X}\} & -10.4\% \\
 How can I become \{\emph{X $\not\eq$ not antonym(X)}\} & -5.5\% \\
 %\midrule
 %traditional SRL: wrong active / passive swap & 2.2\% \\
 %traditional SRL: active / passive swap with people & -6.4\% \\
 %traditional SRL: active / passive swap & -15.2\% \\
%\midrule
 %Change first and last name in one of the questions & -11.5\% \\
 %(q, paraphrase(q)) & -5.3\% \\
\bottomrule
\end{tabular}
\vspace{-5pt}
\caption{
Sample \qqp CheckList tests, with $\Delta$fail\% denoting the failure rate change from \mcomp to \maug. 
%However, the model gets significantly better on \texttt{more X $\not\eq$ less X} by sacrificing \texttt{more X $=$ less antonym(X)}.
With $n=20,000$ and $m=1,911$, \maug failed less on 11 tests (out of the 27 where \mcomp failed), while only becoming worse on 2.
%Meanwhile, the models have similar accuracies on the test set ($84.5 \pm 0.6$ for \maug, and $84.7 \pm 1.0$ for \mcomp).
%Some sample tests are in Table~\ref{table:aug_qqp}.
The model improves consistently on most related cases, but possibly overfits on \texttt{more/less}. 
}
\label{table:aug_qqp}
\vspace{-10pt}
\end{table}}

%%%%%%%%%%%%%%%%%%%%%%%%%%%%%%%%%%%%%%%%%%%%%%%
\section{Counterfactual Evaluation \& Training}
\label{sec:app_label}

We ask crowdworkers to label \sysname-generated counterfactuals for \sst, \nli, and \qqp, for the purposes of evaluation and training.\footnote{We collect \emph{asymmetric counterfactuals}~\cite{garg2019counterfactual} by sampling more \emph{Duplicate} and \emph{Entailment} examples in \qqp and \nli to perturb, due to the difficulty of flipping other labels.} 
In each labeling round, the worker is presented with an original $x$ and its label, and asked to annotate the groundtruth for three $\xp$, rejecting non-fluent ones (details and interface in Appendix~\ref{appendix:label_instruct}).

We use a simple heuristic to select which counterfactuals are presented for labeling, aimed at increasing diversity. 
Representing each $\xp$ by its token changes, control code, and dependency tree structure, we greedily select the ones that are least similar to those already selected for labeling. 
This avoids redundancy in the labeling set, \eg common perturbation patterns such as \swap{black}{white}.

\subsection{Evaluation with Contrast Sets}
\label{subsec:contrast_set}

We verify whether \sysname counterfactuals can be used to create \emph{contrast sets}~\cite{gardner2020contrast}, \ie evaluation sets where each instance has a nearby counterfactual with a \emph{different} groundtruth, to better evaluate model decision boundaries.
We construct these sets by simply filtering out counterfactuals that are labeled the same as their original instances (40\%--63\% depending on the task).

For each task, we test multiple classifers open-sourced by Huggingface~\cite{Wolf2019HuggingFacesTS}, and report the best performing model for each\footnote{\UrlFont{huggingface.co/\{roberta-large-mnli, textattack/roberta-base-SST-2, ji-xin/roberta\_base-QQP-two\_stage\}}} in Table~\ref{table:contrast_set_result} (results for other models are analogous). \sysname contrast sets display performance gaps consistent with those of \citet{gardner2020contrast}, where the sets are constructed manually by NLP researchers, even though we use non-expert annotators who only \emph{label} examples rather than creating them.

\TableContrastSet

\TableAugSST
\TableAugNLI
\TableAugQQP
%%%%%%%%%%%%%%%%%%%%%
\subsection{Training with Counterfactuals}

\label{subsec:augmentation}
%\paragraph{Collection.}
Following \citet{kaushik2019learning}, we augment training sets with counterfactual examples.
% \fixed{and demonstrate \sysname's usefulness by showing that its augmentations produce models that generalize better on \emph{some} datasets without hurting performance on others.}
In all experiments, we finetune \texttt{roberta-base} on datasets of $n$ original examples and $m$ counterfactuals, which are generated by \sysname (\maug) or crafted from scratch by humans (\mcad from \citet{kaushik2019learning}, only available for \nli). To distinguish the benefit of counterfactuals from that of just adding more data, we further add a baseline that uses $n+m$ original examples (\mcomp).
% examples from the original dataset
In addition to in-domain test set accuracy, we measure models' generalization on out-of-domain datasets, as well as contrast sets and challenge sets.
We also evaluate model capabilities with CheckList~\cite{checklist:acl20} for \sst and \qqp.
%As such datasets are unavailable for \qqp, we instead evaluate model capabilities with CheckList~\cite{checklist:acl20}.
Reported model performances are averaged across multiple data samples and random seeds (Appendix~\ref{appendix:data_collection}).

For \sst, we select random \sysname counterfactuals regardless of their labels, as long as an original $x$ has at least one $\xp$ that flips the label.
For \nli and \qqp, we observed in a pilot study that randomly chosen counterfactuals may not be more effective than the same amount of additional data.
We suspect that \sysname lacks domain knowledge and context for identifying critical perturbations, and therefore brings benefits redundant with pre-training~\cite{longpre2020effective}.
Thus, we use the slicing functions of \citet{chen2019slice} to find patterns of interest (\eg prepositions in \nli), and perturb those patterns by placing \texttt{[BLANK]}s on the matched spans. For example, \exinline{His surfboard is beneath him} becomes \exinline{His surfboard is \texttt{[BLANK]} him}, and \sysname generates counterfactuals such as \exinline{His surfboard is \swap{beneath}{next to} him.}

\textbf{Results.}
%Tables \ref{table:aug_sst} \& \ref{table:aug_nli} indicate that \sysname augmentation is effective for \sst and \nli: \maug maintains in-domain test accuracy while consistently improving generalization accuracy in various out-of-domain and challenge datasets. 
%On \nli, \sysname counterfactuals are as effective or more effective than counterfactuals created from scratch (\mcad).
%On \qqp ($n{=}20,000$, $m{=}1,636$), \maug also maintains in-domain accuracy when compared to \mcomp, while significantly lowering error rates (the absolute rate drops for at least $5$ points, with a relative difference of more than $10\%$) in $11$ out of $27$ tests that previously had high error rates (\sysname{} increases error rates in $2/27$ tests).
%\tofix{We aim to demonstrate the usefulness of Polyjuice’s augmentation by showing that it produces models that generalize better on \emph{some} datasets without hurting performance on others. }
Tables \ref{table:aug_sst}--\ref{table:aug_qqp} indicate that \sysname augmentation is effective in all tasks: \maug maintains in-domain accuracy while \emph{consistently} improving or maintaining generalization accuracy in various out-of-domain and challenge sets. 
On \nli, \sysname counterfactuals are as effective or more effective than counterfactuals created from scratch (\mcad).
\fixed{Notably, we obtain the largest gains on challenge and contrast sets (\eg \texttt{Break} and \texttt{DNC} in Table~\ref{table:aug_nli}) or when the out-of-domain dataset is sufficiently different from the training domain (\eg Senti140 and SemEval in Table~\ref{table:aug_sst}). }
\sysname also improves results on CheckList tests that previously had high error rates: 
it significantly lowers the error rates on 11 out of 27 \qqp tests,\footnote{The absolute error rate drops for at least $5$ points, with a relative difference of more than $10\%$.} making $2/27$ tests worse.
For \sst, it improves the model on 5 out of 15 tests, hurting 1.
\fixed{Here, we only report a low $m/n$ ratio (<10\% for NLI and QQP) to show that a small amount of augmentation is already beneficial. 
The results are similar for other combinations we explored (see Appendix~\ref{appendix:data_collection}), except when the ratio of counterfactual to original data was too high (\eg, $m=n$ may decrease vocabulary diversity or induce additional data bias, echoing~\cite{Khashabi2020MoreBF}).}

\subsection{Discussion}
\label{subsec:label_efficiency}
We show that \sysname counterfactuals are useful for evaluation, and more effective than additional (non-counterfactual) data for training in a variety of tasks. 
In contrast to prior work where humans generate counterfactuals from scratch, we only ask them to \emph{label} automatically generated ones, while still achieving similar or better results.

We believe our approach is more effective than manual creation (although both are beneficial): in terms of implementation effort, the process of just labeling counterfactuals is the same as labeling original examples, such that no additional annotator training or separate pipelines are required; in contrast, \citet{kaushik2019learning} set up two separate crowdsourcing tasks for creating and labeling the counterfactuals.
% Further, annotator effort is much lower, as evaluating examples is easier than creating them --- the median time for labeling one round (three $\xp$) was $30$ seconds, while \citet{kaushik2019learning} reported that workers spent ${\approx}4$ minutes per \nli revision (two $\xp$), \emph{prior} to quality validation.
% Even after we remove noisy annotators and labels and disregard non-fluent counterfactuals, we still collect one \nli $\xp$ used in Table~\ref{table:aug_nli} every 0.6 minutes.
Further, annotator effort is much lower, as evaluating examples is easier than creating them --- \citet{kaushik2019learning} report an average of ${\approx}2$ minutes per \nli counterfactual \emph{prior} to quality validation, while our median time was $10$ seconds per counterfactual. Even after our quality validation (removing noisy annotators, disregarding non-fluent counterfactuals), our rate for \nli is ${\approx}36$ seconds per counterfactual (used in Table~\ref{table:aug_nli}).

% the median time for labeling one round (three $\xp$) was $30$ seconds, while \citet{kaushik2019learning} reported that workers spent ${\approx}4$ minutes per \nli revision (two $\xp$), \emph{prior} to quality validation.
% Even after we remove noisy annotators and labels and disregard non-fluent counterfactuals, we still collect one \nli $\xp$ used in Table~\ref{table:aug_nli} every 0.6 minutes.
%1.6 $\xp$ used in Table~\ref{table:aug_nli} per minute.
%\nli revision (two $\xp$, translating to roughly 0.5 $\xp$ per minute), \emph{prior} to quality validation.
%Even after we remove noisy annotators and labels, disregard non-fluent counterfactuals, and apply aggressive post-filtering (\eg only keep $\xp$ that flip labels for contrast sets), we still obtain a suitable \nli $\xp$ in less than one minute.
%one \nli $\xp$ used in Table~\ref{table:aug_nli} every 0.6 minutes.
%we still obtain one suitable \nli $\xp$ used in Table~\ref{table:aug_nli}in 0.6 minute.
%It is also cheaper than labeling non-counterfactuals, as annotators only need to parse the corresponding perturbations, rather than the full instance~\cite{Khashabi2020MoreBF}.

In terms of the utility per counterfactual, manual creation and \sysname may be complementary. 
Manual annotation may be unreliable or incomplete for certain forms of counterfactuals~\cite{ribeiro2018semantically}, whereas \sysname can miss more complex or context-dependent changes, and could benefit from target perturbations that compensate for its lack of domain knowledge (targeted guidance is also helpful for human annotators~\cite{huang2020counterfactually}).
% such as changing the sentence structure
%not always emphasize on critical features if not appropriately controlled.
Thus, it may be important to mix both approaches~\cite{Khashabi2020MoreBF}.
\sysname's flexibility opens up possibilities for hybrids between human creation and human verification of targeted, machine-generated counterfactuals.

\definecolor{cfwone}{HTML}{eef5fa}
\definecolor{cfwtwo}{HTML}{daeaf5}
\definecolor{cfwthree}{HTML}{b2d2e9}
\definecolor{cfwfour}{HTML}{8abbde}

\newcommand{\fwone}[1]{\colbox{cfwone}{#1}\xspace}
\newcommand{\fwtwo}[1]{\colbox{cfwtwo}{#1}\xspace}
\newcommand{\fwthree}[1]{\colbox{cfwthree}{#1}\xspace}
\newcommand{\fwfour}[1]{\colbox{cfwfour}{#1}\xspace}

\newcommand{\fexp}[2]{\texttt{[{\color{darkgray}{#1:#2}}]}\xspace}
\newcommand{\fexptag}[1]{\fexp{TAG}{#1}}
\newcommand{\fexpfrom}[1]{\fexp{FROM}{#1}}
\newcommand{\fexpto}[1]{\fexp{TO}{#1}}
\newcommand{\fexptemp}[1]{\fexp{TEMP}{#1}}

\begin{figure}[t]
\centering
\includegraphics[trim={0 21cm 33cm 0cm},clip,width=1\columnwidth]{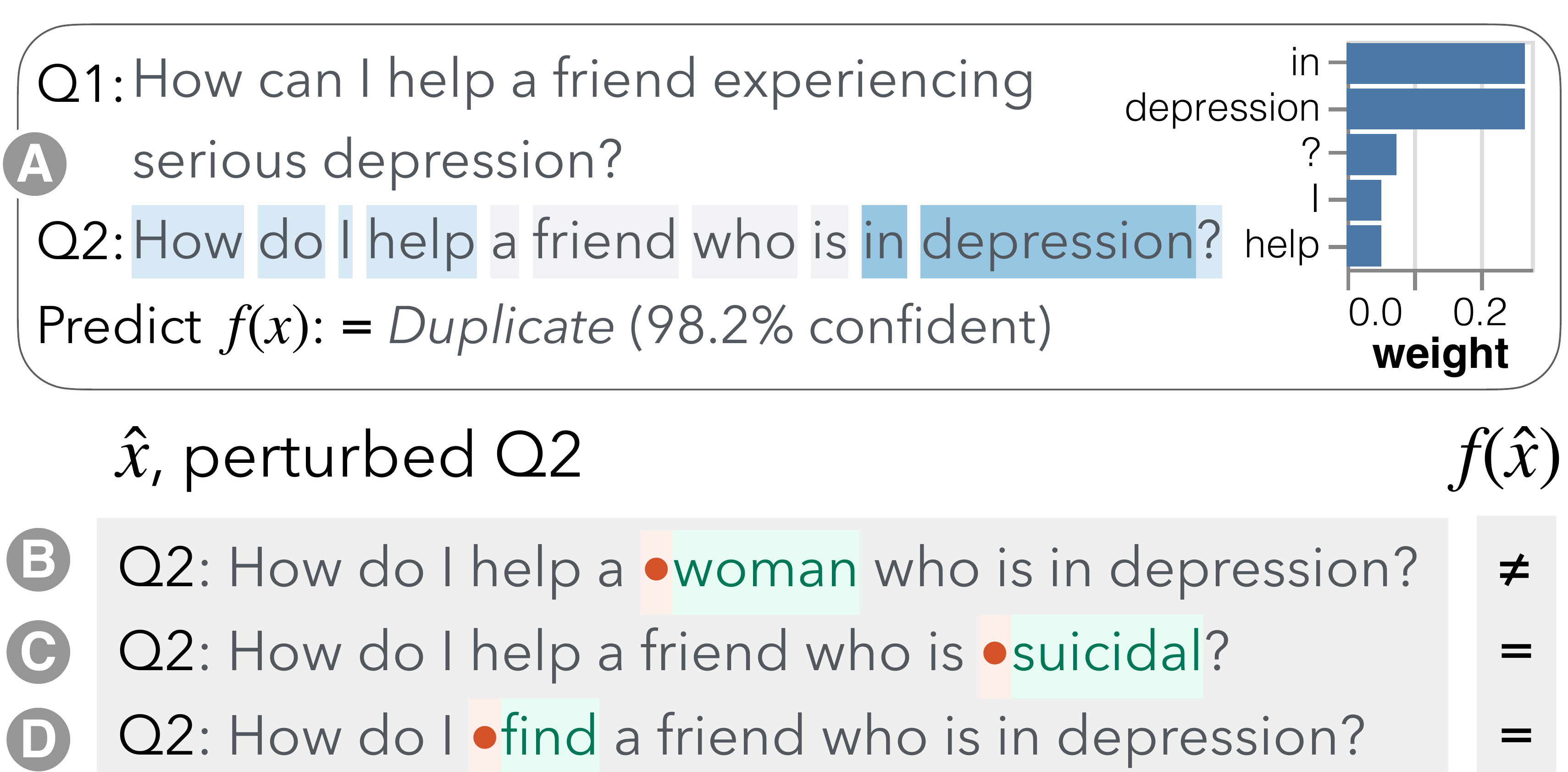}
\vspace{-15pt}
\caption{
(A) An instance in \qqp where the model prediction
$f(x)$ is \emph{Duplicate} ($=$) at 98.2\% confidence, with SHAP importance weights for tokens in Q2.
%Counterfactual explanations complement SHAP with concrete, readable examples, \eg (C) depicts a surprising flipped prediction ($\neq)$ that was missed by SHAP.
Counterfactual explanations complement SHAP with concrete examples and surprising behaviors, \eg (B) shows that \swap{friend}{woman} surprisingly flips the prediction to \emph{Non-Duplicate} ($\neq)$, despite the low weight on ``friend.''
}
\vspace{-5pt}
\label{fig:explanation}
\end{figure}
\section{Counterfactual Explanations}
\label{sec:app_explain}

%Such explanations have been elusive in NLP, despite evidence from social science research~\cite{miller} indicating that they may be more intuitive, or may complement feature attribution or attention maps. 
%Counterfactuals also naturally support model explanations, as ``explanations are sought in response to particular counterfactual cases or foils''~\cite{miller}.
%Popular feature importance attribution methods like SHAP~\cite{NIPS2017_7062} or LIME~\cite{Ribeiro2016WhySI} all retrieve token importance through masking, which can be viewed as a form of (incomplete) counterfactual.

% \subsection{Foils to Feature Attribution Explanations}
A popular way of explaining NLP models is to attribute importance weights to the input tokens, either using attention scores~\cite{wiegreffe2019attention} or by summarizing the model behavior on perturbed instances (\eg LIME~\cite{Ribeiro2016WhySI} and SHAP~\cite{NIPS2017_7062}).
Though ubiquitous, token scores may not always reflect their real importance~\cite{pruthi2020learning}.
Popular packages like LIME or SHAP estimate scores by \emph{masking} words, and therefore may not reflect model behavior on natural counterfactual cases. For example, the token ``friend'' in Figure \ref{fig:explanation}A is not considered important even though a natural substitution in Figure \ref{fig:explanation}B flips the prediction. The opposite happens to ``in depression,'' where a significant change makes no difference to the model's prediction (Figure \ref{fig:explanation}C).
% With the feature importance in Figure~\ref{fig:explanation}A, it is hard for users to imagine the model changing predictions when the trivial token ``friend'' in Q2 is perturbed (in B), or that changing the important ``in depression'' may not matter (in C).
Even perfect importance scores may be too abstract for users to gain real understanding~\cite{miller}, \eg users may not grasp the significance of a low importance score for the token ``help'' without concrete examples such as the one in Figure \ref{fig:explanation}D. 
% --- without the concrete example in Figure~\ref{fig:explanation}D, users may not grasp what it means that the score for ``help'' ${\approx}0.05$. 

% \footnotetext{From BERT \qqp model: \url{https://huggingface.co/textattack/bert-base-uncased-QQP}}

Since presenting a large number of concrete counterfactuals would be overwhelming, we propose a hybrid approach, displaying feature attributions as a high-level summary, together with a judicious selection of \sysname counterfactuals that make behaviors concrete and highlight potential limitations.
Following \citet{miller}'s observation that people look for explanations revealing \emph{unexpected} behavior, we select \emph{surprising} counterfactuals.\footnote{
Details in Appendix~\ref{appendix:exp_rank}.}
That is, we estimate the expected change in prediction with feature attributions, and select counterfactuals that violate these expectations, \ie examples where the \emph{real} change in prediction is large even though importance scores are low (Figure~\ref{fig:explanation}B), and examples where the change is small but importance scores are high (Figure~\ref{fig:explanation}C). 
Of course, users can also view additional counterfactuals that perturb tokens of particular interest, a technique that we explore in the next section.
% Alternatively, based on their own understandings of token scores, \emph{users} can interactively view counterfactuals that perturb certain tokens of interest, a technique we explore in the next section. 

\newcommand{\cshap}{\emph{\sysname-surprise}\xspace}
\newcommand{\crandom}{\emph{\sysname-random}\xspace}
\newcommand{\chuman}{\emph{Expert-surprise}\xspace}

\paragraph{User evaluation.} We study the scenario where an expert has access to a model and local explanations, and evaluate the \emph{additional} benefit of showing counterfactuals, \ie whether they bring \emph{new} insights. 
We evaluate three ways of generating counterfactuals: (1) \crandom, a baseline where we show random \sysname{} counterfactuals, (2) \chuman, where two graduate students (non-participants) were given access to the model and instructed to create counterfactuals that are surprising given the associated SHAP scores, and (3) \cshap, which uses the selection procedure described in the previous paragraph.
%The authors manually checked that all counterfactuals were fluent and unambiguous.
% We aim to answer: does seeing counterfactuals bring new insights, or are counterfactuals redundant with manual analysis or explanations?

We recruited 13 participants (graduate students with experience in model explanation), and had them analyze the aforementioned \qqp model. In each round, users were shown an example, the model prediction, and a SHAP explanation, as in Figure~\ref{fig:explanation}A. Users were instructed to create up to $10$ counterfactuals in order to better understand model behavior around the example, for which model predictions were given (users created $6$ on average). 
Finally, users simulated what the model would do on six counterfactuals~\cite{hase2020evaluating}, two from each condition (in random order). Counterfactuals where users make mistakes are preferable, as displaying these would add information that users do not already have.

\begin{figure}[t]
\centering
\includegraphics[width=1\columnwidth]{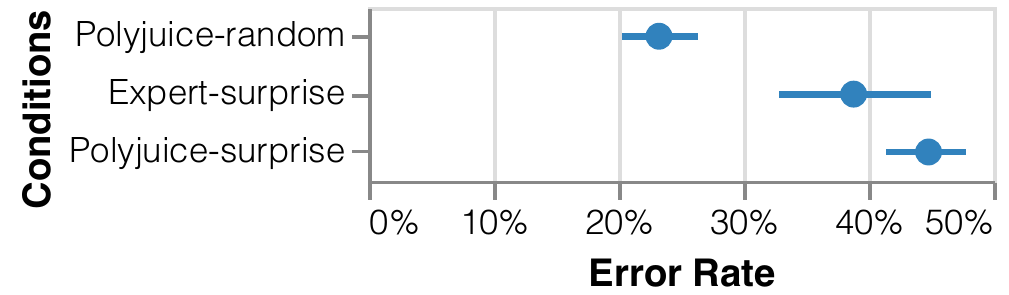}
\vspace{-15pt}
\caption{
Simulation error rates per condition (higher the better). 
\cshap has the highest error rate, indicating these counterfactuals would add the most information to users if displayed.
}
\vspace{-10pt}
\label{fig:err_rate}
\end{figure}
% \end{comment}

%\paragraph{Results.}
As shown in Figure~\ref{fig:err_rate}, humans simulated model behavior on \cshap counterfactuals only slightly better than random guessing ($45\%\pm6\%$), \ie these examples display model behavior that is surprising to users even after seeing explanations and creating their own counterfactuals. \chuman also had a high error rate, but at a much higher cost: generating these for just 20 original instances took 1.5--2 hours of expert labor.

While high error rates could be achieved with unrelated or nonsensical examples, all counterfactuals under evaluation were close to the original examples, when measured by syntactic tree edit (${\approx}1.0$) or Levenshtein distance (${\approx}0.2$), \cshap being the closest on both. An independent rater labeled $95\%$ of \cshap counterfactuals as ``likely written by a native speaker,'' in contrast to $85\%$ for \chuman, indicating that experts sometimes resorted to ungrammatical or nonsensical sentences to find surprising behaviors.

% Have to decide if it's worth keeping this paragraph
Qualitatively, the study participants tended to create counterfactuals by perturbing the token with the highest weights (84\% of their $\xp$ perturbed tokens in the top 15\% quantile of weights), not gaining a real understanding of how the other tokens impact predictions. Participants also made a significant number of mistakes even for tokens they had inspected, \eg a participant perturbed the example in Figure~\ref{fig:explanation}A by replacing \swap{help}{play with}, yielding a \emph{Non-Duplicate} model prediction. When faced with \swap{help}{find} in Figure~\ref{fig:explanation}D, they incorrectly assumed the behavior would be the same.

These results indicate that \sysname{} counterfactuals complement feature attribution explanations by displaying information that users often miss, even after they have manually explored the model behavior \emph{beyond} explanations. Moreover, \sysname{} counterfactuals for this application were more surprising and fluent than \chuman, despite being computed automatically.

\section{Interactive Analysis}
\label{sec:app_err_analysis}

While our use of \sysname has so far relied on automatic selection of counterfactuals, we show in this section how an analyst can benefit from \emph{multiple} counterfactuals per $x$, make use of controlled generation for more advanced analysis, and extract general patterns from individual observations.
Our use case is counterfactual error analysis~\cite{wu2019errudite} of RoBERTa finetuned on \nli (used in \S\ref{subsec:contrast_set}), although the techniques are generally applicable.

\begin{figure}[t]
\centering
\includegraphics[trim={0 12.5cm 33cm 0cm},clip,width=1\columnwidth]{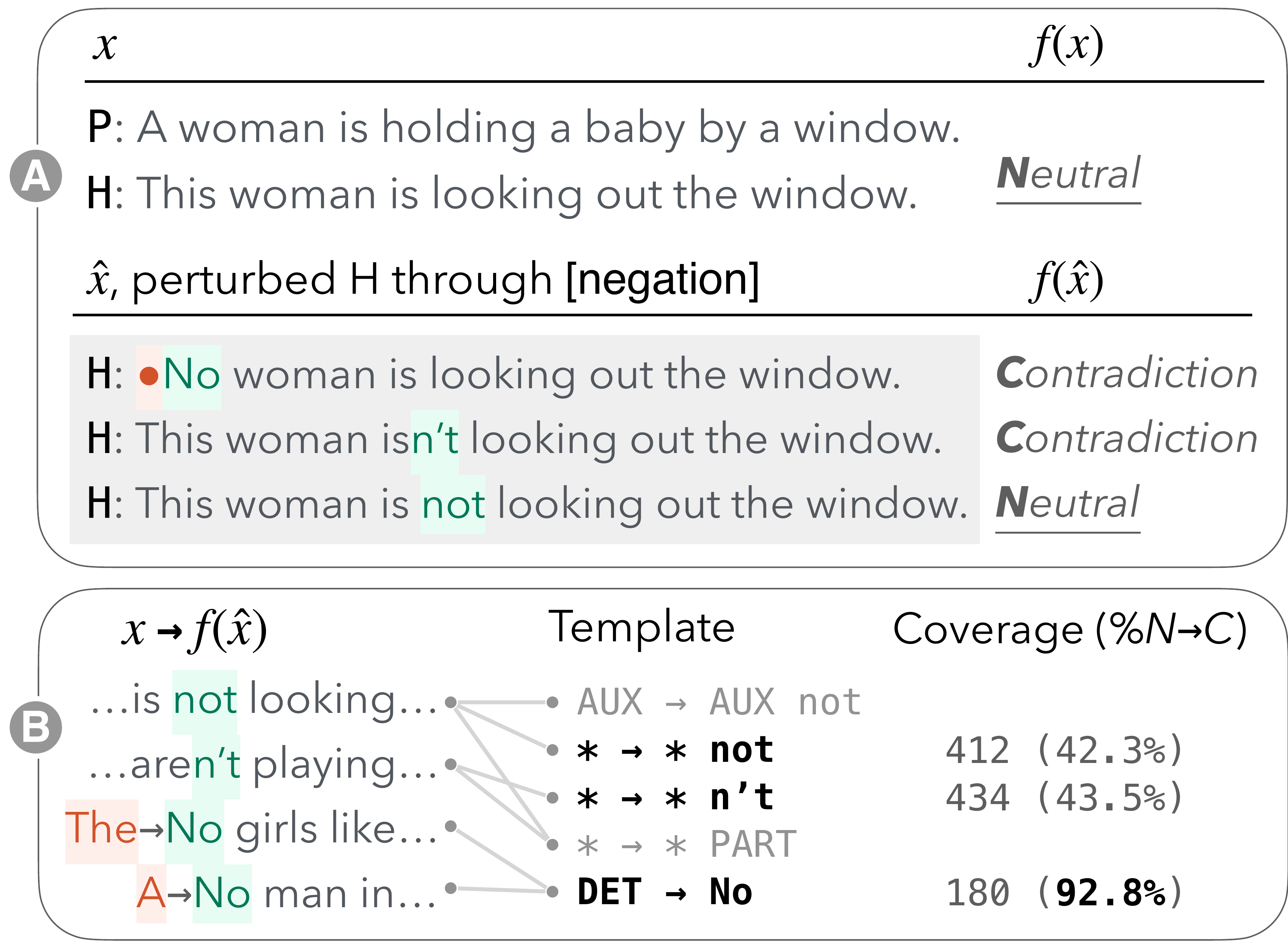}
\vspace{-10pt}
\caption{
(A) An \nli case with a \emph{Neutral} prediction (\uline{underlined} $f(\xp)$ are correct).
\sysname generates counterfactual hypotheses conditioned on the \texttt{negation} \tagstr. 
(B) Generalizing perturbations into patterns~\cite{wu2020tempura}. The change \swap{\texttt{DET}}{no} flips $92.8\%$ of predictions from \emph{N}eutral~$\veryshortarrow$~\emph{C}ontradiction.
%(C) Another blank placement that leads to analyses on \emph{quantifiers}.
}
%\vspace{-5pt}
\label{fig:err_analysis}
\end{figure}

There is a known correlation between the label \emph{Contradiction} and hypotheses with negation in \nli datasets~\cite{gururangan2018annotation}, which may cause models to fail on non-contradiction negations.
We explore this in Figure~\ref{fig:err_analysis}A by generating counterfactual hypotheses for a random \emph{Neutral} instance, conditioning only on the original $x$ and the \texttt{negation} \tagstr.
While the first two counterfactuals display this failure mode, there is a surprising inconsistency in model behavior between ``not'' and ``n't''.
We note that manual analysis may not explore these three negation forms, and thus not surface this puzzling behavior.

\begin{figure}[t]
\centering
\includegraphics[trim={0.5cm 27.5cm 32.5cm 0cm}, clip,width=1\columnwidth]{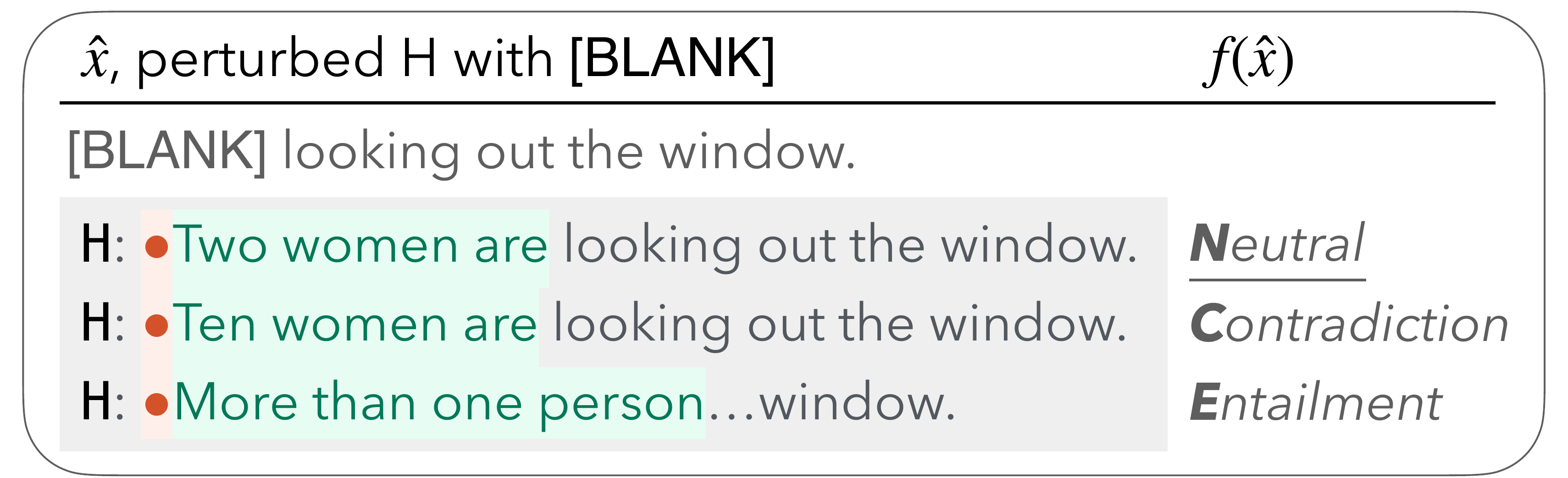}
\vspace{-10pt}
\caption{
Perturbing the subject of $x$ in Figure~\ref{fig:err_analysis}A through \texttt{[BLANK]}, resulting in erroneous predictions for different \emph{quantifiers}
(all should be \uline{\emph{Neutral}}). 
}
%\vspace{-5pt}
\label{fig:err_analysis_quantifier}
\end{figure}

To verify if the pattern is widespread, we generate counterfactuals with the \texttt{negation} \tagstr for a random set of instances correctly predicted as \emph{Neutral} ($n=895$). To generalize individual changes into patterns, we extract frequent \emph{counterfactual templates} with Tempura~\cite{wu2020tempura} (details in Appendix~\ref{appendix:err_analysis_template}), shown in Figure~\ref{fig:err_analysis}B.
The top templates (in bold) show that the model flips its prediction from \emph{Neutral} to \emph{Contradiction} with roughly the same frequency (${\approx}43\%$) whether the negation word is ``not'' or ``n't'', but flips much more frequently with a different negation pattern where a determiner is replaced with ``no'' ($92.8\%$). While these behaviors may be correct in some instances, they often are not (\eg Figure~\ref{fig:err_analysis}A), and thus would warrant further exploration, and potential mitigation strategies (\eg counterfactual training, \S\ref{sec:app_label}).
Tangentially, the impact of \swap{\texttt{DET}}{no} might lead the analyst to explore the impact of perturbing the \emph{subject} of hypotheses, which we do in Figure~\ref{fig:err_analysis_quantifier} by placing a \texttt{[BLANK]} on the subject rather than using a control code.
This leads to the discovery of unstable and erroneous behaviors regarding \emph{quantifiers}, which we analyze in more detail in Appendix~\ref{appendix:err_analysis_quantifier_case}.

\paragraph{Discussion.} 
\sysname{} is a powerful tool for interactive analysis.
Generating multiple counterfactuals per instance leads to insights that might be missed by manual analysis, and the steering provided by \tagstrs and \texttt{[BLANK]}s allow for analyses that would be non-trivial to do manually~\cite{wu2019errudite} or with masked language models (\eg Figure~\ref{fig:err_analysis}B places negations in various parts of sentences, and Figure~\ref{fig:err_analysis_quantifier} replaces spans with other spans of varying lengths). Besides error analysis, an analogous interactive use of \sysname{} may be suitable for test creation~\cite{checklist:acl20} and forms of data augmentation that are more controlled than what we presented in \S\ref{sec:app_label}.

\section{Related Work}
\label{sec:relate}

Some prior work in training and evaluation relies on humans to generate counterfactuals from scratch~\cite{gardner2020contrast, teney2020learning, kaushik2019learning}. 
Our experiments in \S\ref{sec:app_label} indicate that asking humans to \emph{label} \sysname{} counterfactuals yields similar or better results at a lower cost, which motivates an exploration of a mixture of manual and semi-automated generation. 
Similarly, prior work on analysis relies on experts to create individual counterfactuals or perturbation functions~\cite{wu2019errudite, checklist:acl20}. 
In \S\ref{sec:app_err_analysis}, we show that \sysname{} enhances current practice by generating multiple counterfactuals that might have been overlooked, and by providing abstractions that allow for new kinds of analyses.

Prior work on automatically generating counterfactuals typically has a narrower scope in terms of the relationships $x \veryshortarrow \xp$.
For example, adversarial generators aim to maintain semantics while changing model predictions~\cite{ribeiro2018semantically, iyyer2018adversarial, li2020contextualized}, whereas concurrent work to our own~\cite{madaan2020generate, ross2020explaining} automatically generates $\xp$ that change predictions for explanation or analysis, with no constraints on semantics.
However, as shown in \S\ref{sec:app_label}--\S\ref{sec:app_err_analysis}, a \emph{mix} of label-preserving and label-flipping counterfactuals generated by \sysname is quite useful for training, evaluation, explanation, and analysis. 
Further, general-purpose counterfactuals may lead to serendipitous discoveries (\S\ref{sec:app_err_analysis}), especially as \sysname is not fine-tuned to the target domain (and thus less liable to merely replicate what is already there).
Finally, by allowing control through \tagstrs and \texttt{[BLANK]}s, \sysname{} supports human-generator collaboration, where a person specifies desired changes (\eg perturb \emph{the sentence subject}).
Such collaboration is hard to imagine using automatic generators with no control, or with coarser control through predefined style attributes or labels~\cite{madaan-etal-2020-politeness, malmi-etal-2020-unsupervised}. To our knowledge, prior work on controlled generation~\cite{ctrl, pplm} does not address \emph{counterfactual} generation.

\section{Conclusion and Future Work}
\label{sec:discuss}

We propose \sysname, a general-purpose generator that produces fluent and diverse counterfactuals, allowing for control over the kinds and locations of perturbations. 
% With \emph{task-agnostic} controls over perturbation types and locations, \sysname produces general-purpose counterfactuals that are fluent, diverse, close to the original instance.
With simple, \emph{task-specific} selection heuristics, \sysname supports various downstream tasks on different domains, including counterfactual data augmentation, contrast set generation, counterfactual explanation, and error analysis.

While \sysname is broadly applicable, it is not bias-free: control codes are pre-defined and certainly not exhaustive, and the model is fine-tuned on a collection of paired datasets where certain perturbations are more or less likely (\eg we observe that words with negative sentiment tend to be slightly more likely than positive ones in some contexts). 
Collecting naturally occurring counterfactuals is an important area of future research, as is the development of generators that allow for control even without \emph{a-priori} control codes.

\fixed{
Besides improving the generators, further work is needed to improve the value of counterfactuals.
For example, while \sysname shows consistent gains across tasks in data augmentation, the improvements on some datasets are not as significant.
This aligns with observations in prior work that even manual counterfactuals can be marginally beneficial~\cite{kaushik2019learning, huang2020counterfactually}, possibly because the original data is already diverse enough, or the perturbed signal in counterfactuals is too subtle to affect the model (\eg when only a single word is changed in a long sentence.)
%too implicit to grasp.
We hope to perform more thorough experiments on tuning the amount and the distribution of counterfactual augmentation, as well as other ways of incorporating counterfactuals, such as having explicit terms in the loss function for contrasting counterfactuals with original data~\cite{teney2020learning}, or other forms of contrastive learning.
}

\fixed{Although} our applications all involved people, the human-\sysname collaboration in labeling and explanations could benefit from richer interaction mechanisms. 
We believe \sysname motivates future research on more expressive forms of counterfactual training, where users generate counterfactuals together with \sysname, and label counterfactual \emph{patterns} rather than individual instances. 
Similarly, interactive explanations and analysis are exciting directions, especially as we develop new ways of selecting, presenting, and aggregating counterfactuals for various analysis objectives.
%to coordinate with various analysis objectives of humans.
%aggregating counterfactuals and their effects. 
Having noted these opportunities, we believe \sysname is already a powerful tool for counterfactual reasoning, in particular for tasks where people are directly involved. 
\sysname is opensource, and available at \modelurl.

\section*{Acknowledgements}
The work was supported by ONR grant N00014-18-1-2193, NSF RAPID grant 2040196, NSF award IIS-1901386, the University of Washington WRF/Cable Professorship, and the Allen Institute for Artificial Intelligence (AI2).
We thank 
Jim Chen, 
Dianqi Li,
Scott Lundberg, 
Hao Peng, 
Sameer Singh,
Jiao Sun,
Victor Zhong,
and Sitong Zhou for their helpful comments, as well as our user study participants for their valuable input.
%We also appreciate the valuable input from our user study participants.

%\clearpage
\newpage

\section*{Ethical Considerations}
Our work includes labeling counterfactuals on crowdsourcing platforms, as well as conducting user studies with graduate students.
As detailed in Appendix~\ref{appendix:label_instruct} and \ref{appendix:exp_user_study}, we compensated the MTurk workers \$2.5 for ${\approx}15$ minutes of labeling, and the graduate students \$20 for the user study (one hour), above the U.S. federal minimum wage.
The studies are with IRB approval.

We only finetune GPT-2 rather than training it from scratch, such that our compute costs are relatively low (around 8 hours for finetuning, Appendix~\ref{appendix:train_data}). All of our finetuning experiments involved finetuning RoBERTa on smaller datasets.
% To reduce the cost, we may consider optimizing the finetuning dataset to reduce repetitive sentence pairs, and thereby train \sysname with fewer shots.

More critically, with most of our demonstrated applications using a human-generator hybrid mechanism, we stress that the interaction between the two deserves careful consideration.
It has long been reported that algorithms interacting with humans can negatively impact the human.\footnote{\url{https://www.nytimes.com/interactive/2017/04/02/technology/uber-drivers-psychological-tricks.html?_r=0}}
In our case, the concern might be that users can develop an over-reliance on \sysname~\cite{bansal2021does} and hastily accept its generations.
Not only can this decrease users' creativity~\cite{green-etal-2014-human}, but it may bias their analysis process: as discussed in \S\ref{sec:discuss}, \sysname generation is not exhaustive, and may favor some perturbation patterns over others in unpredictable ways.
% Hastily relying on \sysname generations may risk incomplete explorations.
In the short term, we plan to highlight these limitations as part of the model documentation, while future research should identify interaction mechanisms, so as to ensure that \sysname or other counterfactual generators support humans, rather than hindering their performance.

\bibliography{ref}
\bibliographystyle{acl_natbib}

\clearpage
\newpage

\appendix
%\begin{comment}
\begin{table*}[t]
\small
\centering
\setlength{\tabcolsep}{4pt}
\begin{tabular}{@{}lrrrrrrrrr@{}}
\toprule
\textbf{Dataset} & \textbf{\ctrltag{negation}} & \textbf{\ctrltag{quantifier}} & \textbf{\ctrltag{lexical}} & \textbf{\ctrltag{resemantic}} & \textbf{\ctrltag{insert}} & \textbf{\ctrltag{delete}} & \textbf{\ctrltag{restructure}} & \textbf{\ctrltag{shuffle}} & \emph{\ctrltag{global}} \\ 
\midrule
        CAD &      3,274 &         292 &    8,143 &         2,603 &      960 &         952 &          220 &       36 &    3,466 \\
   Contrast &       336 &         436 &     1,607 &         1,291 &      589 &         586 &          275 &      149 &     877 \\
       HANS &        50 &           0 &        0 &              0 &      3,926 &       3,926 &          494 &     1,602 &       2 \\
    ParaNMT &      2,797 &         825 &    10,000 &        10000 &    6,442 &       6,205 &         5,136 &     1,417 &   10,000 \\
       PAWS &        81 &        1,815 &    10,000 &        10000 &    3,630 &       3,403 &         4,551 &    10,000 &   10,000 \\
 WinoGrande &      3,011 &          94 &    10,000 &        6,927 &     120 &         124 &          453 &       65 &    3184 \\
    \emph{Crawled} &         0 &           0 &     5,000 &           0 &    5,000 &    5,000 &            0 &      108 &    5,000 \\
      \textbf{Total} &      9,549 &        3,462 &    44,750 &       30,821 &   20,667 &   20,167 &        11,129 &    13,377 &   32,529 \\
\bottomrule
\vspace{-15pt}
\end{tabular}
\caption{The datasets used for finetuning \sysname, and the \tagstr distributions.}
\label{table:gpt_train_stats}
\vspace{-10pt}
\end{table*}

\section{GPT-2 as Counterfactual Generator}
\label{appendix:train_data}

\subsection{Training Data and Parameters}

We combine several datasets to finetune \sysname.

\textbf{Contrast set.}
Authors of 10 existing NLP dataset each manually perturbed 100--1,000 instances to change the gold label, so to inspect a model's local decision boundary~\cite{gardner2020contrast}.
The perturbation patterns vary based on the tasks and the annotators, allowing us to learn diverse strategies.
To make sure we can use the contrast set to evaluate the \sst model, we excluded the IMDb movie review from the training.
%\wts{Re-train the model with other contrast sets.}

\textbf{Counterfactually-augmented data (CAD).}
\citet{kaushik2019learning} crowdsourced counterfactuals for IMDb movie review (1.7k), which we split into paired sentences to match the text length of other datasets.
CAD's perturbation patterns also vary based on the task, but can especially contribute to \ctrltag{negation}.
As NLI is in our demonstrating applications, we did not use their 6.6k SNLI counterfactuals.\footnote{Similarly, though \qqp is suitable for training \sysname, we omitted it so \qqp can be used in our evaluation.}
% and SNLI (6.6k).

\textbf{WinoGrande} is a large-scale dataset of 44k instances for testing common sense problems~\cite{sakaguchi2019winogrande}.
It contains sentences that differ only by one trigger word (\eg one noun), making it most suitable for learning lexical exchanges.

\textbf{ParaNMT-50M} contains 50 million English-English sentential paraphrase pairs, covering various domains and styles of text, as well as different sentence structures~\cite{wieting2017paranmt}. 

\textbf{PAWS}~\cite{zhang2019paws} contains pairs with high text overlaps, created through controlled word swapping, best demonstrating \ctrltag{shuffle} and \ctrltag{restructure}. We used its 49k Wikipedia parts.

\textbf{HANS}~\cite{mccoy2019right}, a challenge set for NLI, contains 10k pairs of premises and hypotheses created based on 10 heavily fallible syntactic templates, and therefore compensates rarer structural changes that may be missed by PAWS.

\textbf{Crawled} 
We additionally crawl naturally occurring sentence pairs from non-paired datasets boost some specific patterns and increase lexical diversity. 
This include 
(1) CommonGen~\cite{lin-etal-2020-commongen}, sentences with common sense concepts; 
(2) Natural Questions~\cite{kwiatkowski-etal-2019-natural}, collections of queries issued to Google Engines (and therefore involve various paraphrases of similar user intents), and 
(3) SQuAD \cite{rajpurkar-etal-2016-squad}, whose paragraphs involve Wikipedia knowledge.
We estimate \emph{close} pairs using edit distance, and broadly accept those with less than 60\% editing.
To exclude tricky cases (\eg \exinline{how do I not be} can be incorrectly regarded as \ctrltag{negation} for \exinline{how do I recover it}), we only augment the most determined patterns: \ctrltag{lexical}, \ctrltag{insert}, \ctrltag{delete}, and \ctrltag{shuffle}.

\fixed{To balance the distribution (Table~\ref{table:gpt_train_stats}), for each dataset, we extract \tagstrs from all the $(x, \xp)$, and randomly sample up to 10,000 instances per \tagstrshorts.
Still, \ctrltag{quantifier} and \ctrltag{negation} have less training data compared to other codes. 
Fortunately, these codes tend to be limited to more specific patterns (``more than'', ``not'', ``never'') when compared to ``broad'' codes like \ctrltag{lexical}, and thus even a small sample is enough to learn them. }
We finetuned an off-the-shelf GPT-2 model from \citet{Wolf2019HuggingFacesTS} for 10 epochs with an initial learning rate 5e-5, a batch size of 8, and a sequence length of 120 (but any LM can potentially be used).
We select the best epoch based on the evaluation loss on a holdout set of size 5,000.
The training took around 8 hours on two Titan RTXs.

\subsection{Intrinsic Evaluation Details}
\label{appendix:intrinsic}

\subsubsection{Closeness and Diversity}
\label{appendix:closeness}
Similar to \citet{madaan2020generate}, we compare the \emph{diversity} and \emph{closeness} of \sysname with alternative generators, \ie RoBERTa and T5, representing masked language models that prioritize word and span substitution, and original \emph{GPT-2}, representing the standard generative model not conditioned on $x$. 
For a given $x$ and its counterfactuals $\hat{\xset}$, we approximate \emph{diversity} using self-BLEU~\cite{zhu2018texygen} within $\hat{\xset}$.
Meanwhile, \emph{closeness} is the average distance between $x$ and every $\xp \in \hat{\xset}$, both with the normalized word level Levenshtein edit distance (\cite{levenshtein1966binary}, used in MiCE~\cite{ross2020explaining}), and syntactic tree edit distance (\cite{zhang1989simple} in GYC~\cite{madaan2020generate}).

We run the three generators on 300 sentences in total.
In GPT-2, we take the first two words of an $x$ as the input context (prompt), limit the length of the generation to be similar to $x$, and collect 10 counterfactuals.
As for RoBERTa and T5, we repeatedly perturb $x$ for three times, each time randomly placing up to three \texttt{[MASK]} tokens, and ask the generator to generate 5 counterfactuals through beam search, following \citet{checklist:acl20}.
\sysname uses the same blank (mask) placement as in RoBERTa and T5, but we additionally enumerate through all \tagstrs.
For each $x$, we randomly sample 5 counterfactuals to form $\hat{\xset}$ per generator.

As shown in Table~\ref{table:intrinsic}, \sysname achieves a balance between diversity and closeness.
Ideally, we would also like to compare \sysname with concurrent work \cite{madaan2020generate, ross2020explaining}, but these are yet to be open-sourced and require extensive implementation or finetuning.
%Inspired by style transfer~\cite{yang2018unsupervised} and controlled text generation, GYC performs the perturbation on the latent space of the input $x$.
%Meanwhile, MiCE uses a two-step framework to generate counterfactual explanations, with the generator being T5~\cite{JMLR:v21:20-074} finetuned on the task-specific dataset.
%As mentioned in \S\ref{sec:relate}, both generators focus on flipping the class label of a given $x$.
% Unfortunately, both require extensive implementation or finetuning, and has yet to be opensourced.

%Still, we believe that our infilling (\texttt{BLANK}) structure will improve the \emph{closeness} over the unconstrained GYC (called content and syntactic preservation in their case).
%We also hypothesize that \sysname has better (or at least compatible) diversity than GYC. 
%Per their intended use case --- model testing and debiasing --- the control functions in GYC are driven by the predictor-of-interest (\eg sentiment classifier, NER tagger). 
%As a result, most of the GYC changes seem to focus on features deemed important by the classifier (\eg \exinline{I am very disappointed with the service} is changed by \swap{disappointed}{pleased}, \swap{disappointed}{happy}, \swap{disappointed with}{pleased to get a good}).

\subsubsection{Controllability}
\label{appendix:ablation_control}

\fixed{To evaluate controllability, we compare \sysname with T5, and GPT-2 finetuned on prompts \emph{without} \tagstrshorts (called \sysname\emph{-a}), such that both baselines consider sufficient context.}
%We finetune GPT-2 on training prompts that \emph{do not} contain \tagstrs (called \emph{\sysname-a}), and quantify the impact of the \tagstrshorts through an ablation study.
For each \tagstr, we compare the \emph{control success rate} of \sysname and \sysname-a on 300 prompts.
For each prompt, we generate counterfactuals through beam search (beam $=5$), and recompute the \tagstrshorts on the top three generated $\xp$.
We deem the control successful if at least one of the three recomputed \tagstrshorts matches the desired control code (though in \sysname-a, we only measure whether the \tagstrshort naturally occurs in the uncontrolled generation.)
The success rate increases by $26\% \pm 13\%$ across all \tagstrs, ranging from \ctrltag{quantifier} (increasing 6\%, from 50\% to 56\%) to \ctrltag{negation} (42\%, from 5\% to 47\%).
\fixed{Non-finetuned T5 also achieves less control (success rate decreases by 33\% on average.)}

%\ctrltag{lexical} has the smallest increment \tofix{from $93\%$ to $100\%$}, mostly because \emph{\sysname-a} tend to frequently replace words.
%\ctrltag{insert} is the most impactful \tagstrshort (from \tofix{$12\%$ to $100\%$}) --- \sysname-a rarely insert additional clues on its own.

Common failure cases include
%Note that all the prompts used are guaranteed to allow the corresponding \tagstrs, and the \tagstrshorts can be less effective on more general prompts:
(1) The \tagstrs conflict with the blanks, \eg \exinline{a dog is embraced by a \texttt{[BLANK]}} would not respond to \ctrltag{negation}.
(2) $x$ does not have a corresponding pattern, \eg \ctrltag{shuffle} is not applicable to \exinline{the movie is good.}
(3) certain salient patterns dominate the generation probability, \eg the model tends to perturb the quantifier ``two'' in \exinline{two dogs are running,} regardless of the \tagstrshort.
%In the ablation study, we filtered out prompts that fell under cases 1 and 2.

\begin{comment}
{'gpt2': {'bleu4': 0.17553428639447374,
  'bluescore': 0.8947643,
  'sem_dist': 0.6225155562821245,
  'tree_dist': 6.350782997762864,
  'edit_dist': 0.7045303072049199},
 'bert': {'bleu4': 0.4699445059841606,
  'bluescore': 0.9546045,
  'sem_dist': 0.15153456281610847,
  'tree_dist': 1.352,
  'edit_dist': 0.14362360165631063},
 'polyjuice': {'bleu4': 0.33819550232972273,
  'bluescore': 0.9369372,
  'sem_dist': 0.23683031172394833,
  'tree_dist': 2.1298870056497172,
  'edit_dist': 0.2505382626667885}}
\end{comment}

%\input{sections/app_survey}

\begin{comment}
\begin{figure*}
\centering
\includegraphics[width=0.9\textwidth]{figures/mturk_instruction.pdf}
\vspace{-1pt}
\caption{
The instruction for the \nli labeling task in \S\ref{sec:app_label}, with annotators labeling the perturbed hypotheses (\emph{New S2}). 
Instructions are similar for \qqp and \sst, except for the label definitions and the examples.
\wts{Maybe remove.}
}
\vspace{-10pt}
\label{fig:mturk_instruction}

\end{figure*}
\end{comment}

\section{Additional Train \& Eval Details, \S\ref{sec:app_label}}
\label{appendix:app_label}

\begin{figure}
\centering
\vspace{10pt}
\includegraphics[width=0.9\columnwidth]{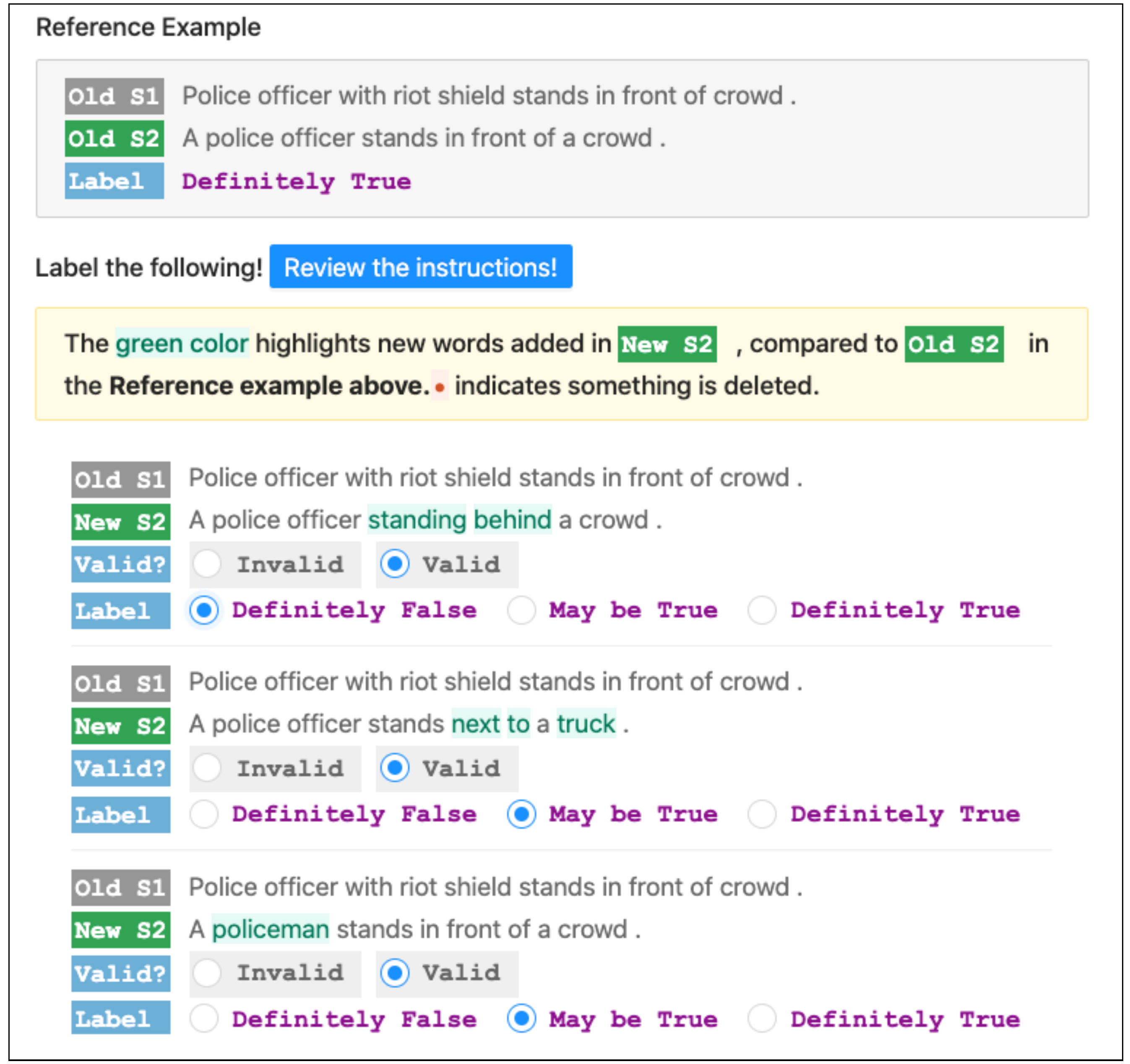}
\vspace{-5pt}
\caption{A sample labeling task: The crowdworkers annotate three counterfactuals based on their validity and class label, with respect to the original instance.}
\vspace{-15pt}
\label{fig:mturk_ui}
\end{figure}

\subsection{MTurk Labeling Details}
\label{appendix:label_instruct}

\textbf{Procedure}
%, in which we explained the context and tasks (Figure~\ref{fig:mturk_instruction})
The study started with an introduction that explained the context and tasks.
%: given a reference example, the crowdworker should annotate its counterfactual variations, based on whether the counterfactual is valid (\emph{fluent}), and the classification task label.
To familiarize crowdworkers with the task, we asked them to complete 1-2 training rounds, and explained the expected labels.
Each annotator then completed 22 tasks, labeling 3 counterfactuals of a single example in each round, as in Figure~\ref{fig:mturk_ui}.
The 22 rounds consisted of 20 actual labeling tasks and 2 extra ``gold rounds'' with known correct labels.
The gold cases later served to filter low-quality crowdworkers.
%As a result, each annotator contributed $20 \times 3=60$ labels.
% (14.9 for \qqp, 16.7 for \sst, and 19.8 for \nli)
The median annotation time was around 15 minutes, and participants received \$2.5.

\textbf{Participants.}
We recruited participants from MTurk, limiting the pool to subjects from within the US with a prior task approval rating of at least 97\% and a minimum of 1,000 approved tasks.

\textbf{Data quality.}
We applied two filtering strategies: 
(1) \emph{High-quality worker.} 
We only kept data from participants whose median labeling time per round was more than 18 seconds and correctly labeled at least 4 gold counterfactuals (out of 6), or who correctly labeled all gold ones.
(2) \emph{Majority vote labeling.}
We collected two annotations per counterfactual, and only kept those that at least one annotator deemed valid, and both annotators agreed on a particular class label.
%Due to the crowdsourcing noise, when set out to collect counterfactuals on 1,000 original examples (thus 3,000 counterfactuals), we typically collect counterfactuals for 1,000 counterfactuals on 600 original examples.
One of the authors labeled a subset of 100 $\xp$ on 100 $x$ in \sst, and reached high agreement with the majority-voted results ($\kappa=0.77$, raw labeling agreement $88\%$).

\subsection{Training Details \& $m/n$ Ratios, for \S\ref{subsec:augmentation}}
\label{appendix:data_collection}

%\paragraph{Model.}
%We finetuned \texttt{roberta-base} models~\cite{liu2019roberta} provided by HuggingFace Transformers~\cite{Wolf2019HuggingFacesTS}.
For each $(m,n)$, we created three samples of training data.
Each sample was further averaged over four random seeds.
For each run, we heuristically picked the initial learning rates 1e-5, 2e-5, 2e-5 for \sst, \nli and \qqp, and trained 20 epochs with a dropout rate of 0.1 and a batch size of 16. 
We selected the epoch that had the highest accuracy on the corresponding validation set, which takes 1/5 of the training data size, with the same ratio of $m/n$ counterfactual and original examples.
%\wts{Double check the reproducibility requirement to see if there's anything missing. And should I move the whole section to appendix?}

\fixed{
\begin{figure}[t]
\centering
\includegraphics[width=0.95\columnwidth]{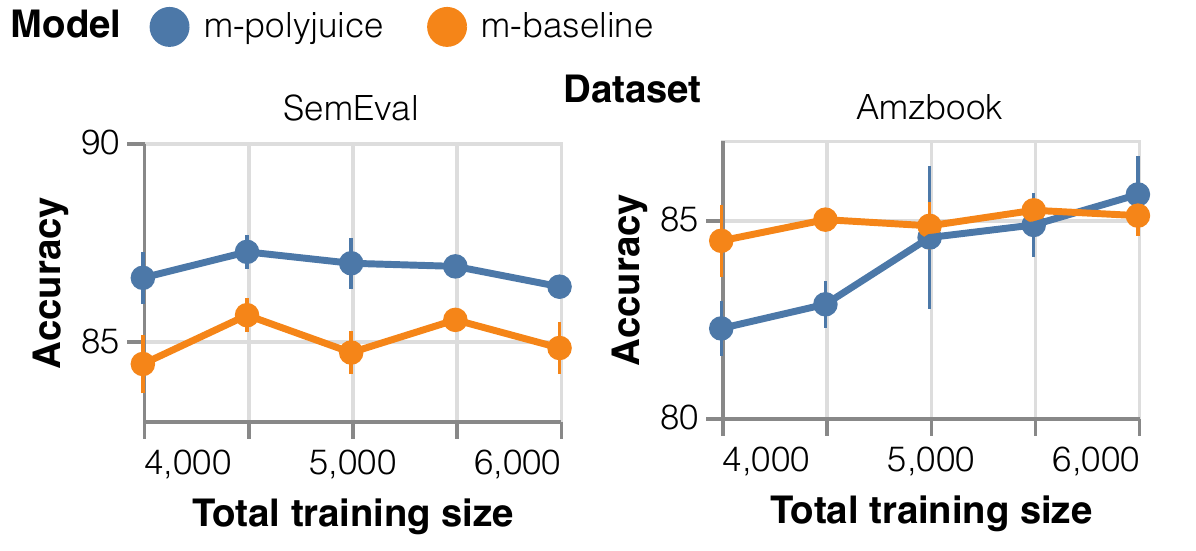}
\vspace{-5pt}
\caption{
\fixed{The accuracy trend on two \sst datasets, as the total training datasize ($m+n$) varies. The blue line shows an augmentation of $m=2k$ counterfactuals, and the blue one represents the corresponding \mcomp.
Though the counterfactuals remains useful on datasets like SemEval across all $m+n$, it appears too many counterfactuals may be harmful (Amzbook).
}
% (when $m+n=4k$, we have $m=n=2k$ on the orange line)
}
\vspace{-10pt}
\label{fig:sst_trend}
\end{figure}

%\emph{The ratio of the counterfactuals matters. }
%In both \nli and \qqp, just adding $m/n < 10\%$ data is sufficient to boost the performance.
We further explore ratios of added counterfactuals. 
%The ratio of counterfactual also matters.
Take \sst as an example: while the counterfactual remains effective on most datasets, it hurts the model performance on Amzbook when the counterfactual takes a large proportion (Figure~\ref{fig:sst_trend}, Yelp followed a similar but more mild trend).
We suspect that flipping out too much original data affects the data diversity, and in turn decreases the model performance.
Similarly, \citet{huang2020counterfactually} asserted that augmenting $n=1.7k$ NLI data with $m=6.6k$ counterfactuals did not improve model generalization accuracy.% over augmenting with non-counterfactual data.
}

\begin{comment}
\subsection{Criteria for CheckList Tests}
\label{appendix:checklist}
\wts{Maybe remove.}
As a behavioral testing framework, CheckList defines multiple tests and measures models' linguistic capabilities using the failure rate of each test.
We define a test to have failed if the failure rate is over $20\%$.
Because failure rates are more sensitive than the accuracy, we say a model capability is affected, if the failure rate of a test changes (increases or decreases) more then 5\% (\eg failure rates going from 20\% to 21\% is insignificant), and the delta accounts for 10\% (\eg failure rate decreasing 8\% from 100\% to 92\% does not count.)
\end{comment}

\begin{comment}
\begin{figure*}[ht]
\centering
\includegraphics[width=1\textwidth]{figures/explanation_instruction}
\vspace{-15pt}
\caption{The instruction for the explanation study in \S\ref{subsec:exp_user_study}. \wts{Maybe remove.}}
\vspace{-10pt}
\label{fig:explanation_instruction}
\end{figure*}
\end{comment}

\begin{figure*}[ht]
\centering
\includegraphics[width=0.81\textwidth]{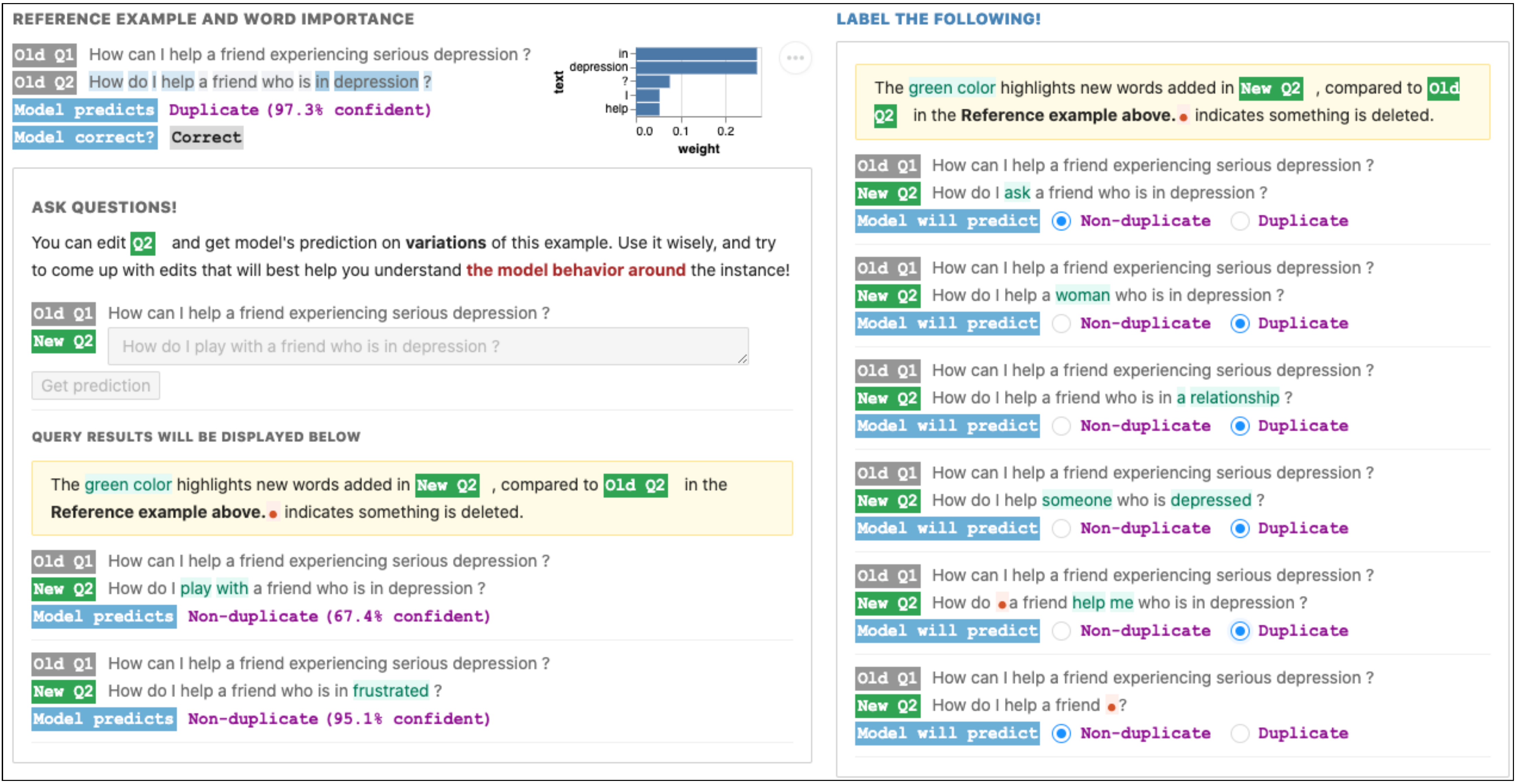}
\vspace{-5pt}
\caption{A sample explanation task for \S\ref{sec:app_explain}}
\vspace{-10pt}
\label{fig:explanation_ui}
\end{figure*}

\section{Additional Explanation Details \S\ref{sec:app_explain}}
\label{appendix:explanation}

\subsection{Selection Methods}
\label{appendix:exp_rank}

Because  SHAP weights reflect the average effect of masking a token $t$, we also focus on \emph{word features that are abnormal on average}.

More concretely, we define the expected change-in-prediction for perturbing a token $t$ to be the SHAP importance on it, $\E[\dist(t, x)] = s(t)$.
In Figure~\ref{fig:explanation}, $s(t\text{=\remove{depression}})=0.276$.
The actual prediction change $\dist(t, x)$ is the weighted average of $|\fp(x)-\fp(\xp)|$ for all the $\xp$ that affect $t$ (\swap{depression}{trouble}, \swap{depression}{a mood}), where $\fp(x)$ is the prediction probability of $f$ on $x$.
The weight corresponds to the number of words modified in $\xp$: If $e(\xp)$ denotes the set of edited words in $x$, then $w(\xp) = 1/|e(\xp)|$.
Intuitively, the more words changed in $\xp$, the less impact each word has; In Figure~\ref{fig:explanation}D, we regard ``depression'' to be responsible for half of the impact in \swap{in depression}{suicidal}.
We group $\xp$ based on their affected words $G_t = \{\xp\ |\ t \in e(\xp)\}$. $\dist(t, x)$ then becomes:
$$\dist(t, x) = \frac{1}{|G_t|+1} \left(s(t) + \sum_{\xp \in G_t} w(t)\cdot |\fp(x)-\fp(\xp)|\right)$$
The additional SHAP weight $s(t)$ acts as a smoothing factor to penalize outliers.
Then the gap between the expectation and reality is:
$$\Delta\dist(t, x) = \dist(t, x)-\E[\dist(t, x)]$$
We first find the abnormal tokens: (1) $t$ with small SHAP weight, but $\xp$ that change $t$ experience large prediction change on average: $t_L = \argmax_{t\in x} \Delta\dist(t, x)$, and (2) $t$ with large SHAP weight, but $\xp$ with $t$ changed usually have intact prediction: $t_U = \argmax_{t\in x} -\Delta\dist(t, x)$.

Then, we use the most extreme cases within the groups of $G_{t_L}$ and $G_{t_U}$ as the concrete counterfactual explanations, based on their prediction change $|\fp(x)-\fp(\xp)|$, and the aggregated SHAP weights of all the changed tokens:
$$\xp_L = \argmax_{\xp \in G_{t_L}} \left( |f_p(x)-f_p(\xp)| - \sum_{u\in r(\xp)} s(u) \right)$$

\subsection{User Study Details}
\label{appendix:exp_user_study}

%The instruction for the user study in \S\ref{subsec:exp_user_study} is in Figure~\ref{fig:explanation_instruction}, and 
Figure~\ref{fig:explanation_ui} shows the sample interface. 
Participants started by just seeing the reference example and the model query box on the left hand side.
When they chose to start the task or after they had exhausted their ten query chances, the query box was disabled, the tasks on the right were displayed, and the participants completed the tasks.
We compensated participants \$20 for the one hour study.

\section{Additional Err. Analysis Details \S\ref{sec:app_err_analysis}}
\label{appendix:err_analysis}

\subsection{Additional Case Study: Quantifiers}
\label{appendix:err_analysis_quantifier_case}

\begin{figure}[t]
\centering
\includegraphics[trim={0 25.2cm 34.5cm 0cm},clip,width=1\columnwidth]{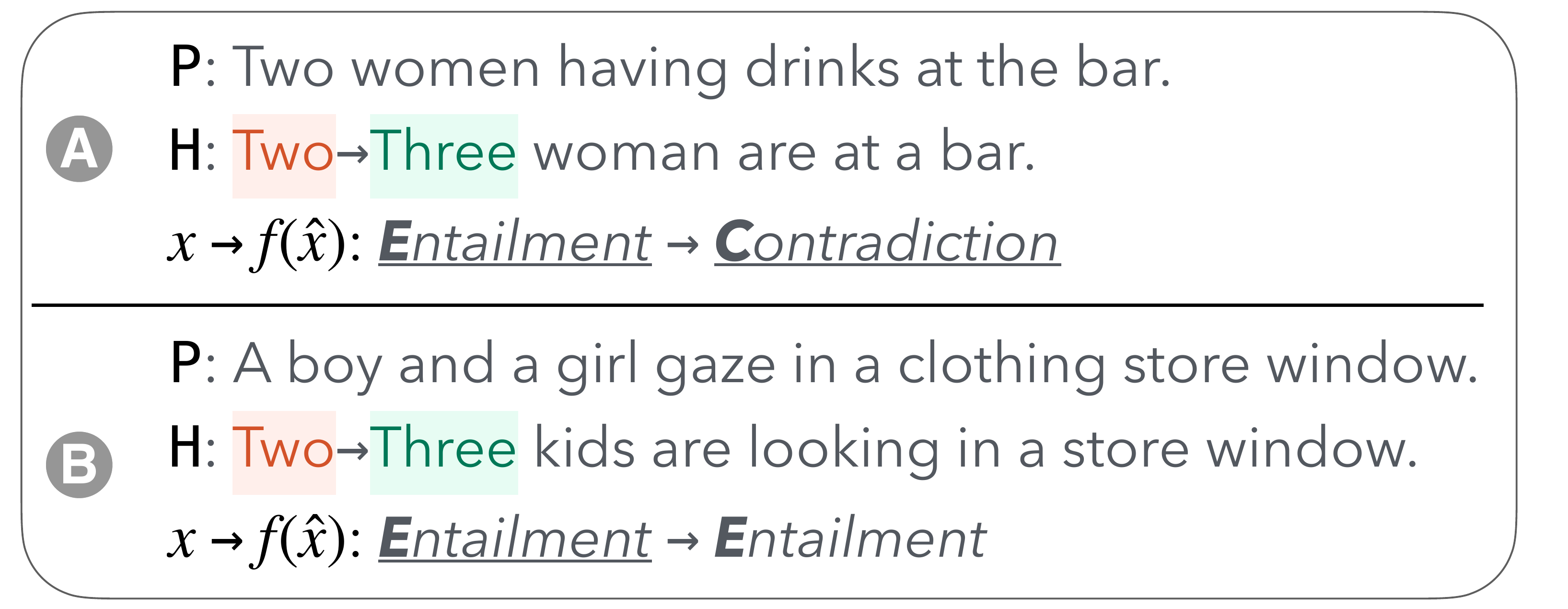}
\vspace{-20pt}
\caption{
The \nli model cannot perform the actual counting when the exact number is missing from \emph{P}.
%In (B), it still predicts (the now incorrect) \emph{Entailment}.
}
\vspace{-15pt}
\label{fig:err_analysis_two_three}
\end{figure}

%\citet{gururangan2018annotation} also mentioned that one common strategy for creating NLI examples is to modify the numbers, suggesting that the model should have seen various examples related to numbers.

As a follow-up to Figure~\ref{fig:err_analysis_quantifier}, we slice the data to find \emph{entailment} instances that have numbers in the hypothesis sentence, and perturb their \ctrltag{quantifiers}.
The extracted templates show that the model does not perform actual counting. 
When changing one number to another (\swap{\texttt{NUM}}{\texttt{NUM}}), the model only flips the label in 64.7\% cases, while we would expect all cases to be like in Figure~\ref{fig:err_analysis_two_three}A.
An inspection of instances indicates the model gets confused when the premise does not contain the same number explicitly. 
Indeed, when we filter for such instances (e.g. Figure~\ref{fig:err_analysis_two_three}B), the label flip rate of \swap{\texttt{NUM}}{\texttt{NUM}} is lowered to 30.2\%.

Further, the model only reacts to \emph{some} quantifier phrase modifiers. 
\addplus{at least} (\exinline{\add{at least} two women are at a bar}) will always still result in \emph{entailment}, prediction, \addplus{only} and \addplus{exactly} flip the predicted label to \emph{neutral} 90\% of the time (\exinline{\add{exactly} two women are at a bar}), but the model only changes the prediction 52.6\% of the time when we add \addplus{more than} (\exinline{\add{more than} two women are at a bar}).

\subsection{Representative Perturbation Templates}
\label{appendix:err_analysis_template}

Similar to \citet{wu2020tempura}, the process of finding representative perturbation patterns takes two steps:

\textbf{Extract template.}
For each $\xp$, we compare it with its $x$, and translate the perturbed spans into templates using different combinations of texts, lemmas, sparse and fine-grained part-of-speech tags.
We optionally include surrounding contexts determined by the dependency tree structure (tokens that share the same parents as the perturbed span).
For example, \exinline{is \add{not} reading} can result in templates $t$ as fine-grained as \swap{is reading}{is not reading}, or as sparse as \addplus{\texttt{PART}}.
Meanwhile, \exinline{are \add{not} playing} also translates to \addplus{\texttt{PART}} or \addplus{\texttt{not}}, but not \swap{is reading}{is not reading}.
As such, the $\xp$ and templates form a many-to-many relationship: each $\xp$ generates multiple templates, and each template covers a different group of $\xp$.

\textbf{Select Representative Templates.}
To find representative changes, we prefer (1) templates that cover a large number of $\xp$.
Meanwhile, to avoid overfitting to one instance (\eg extracting a template \swap{red}{\texttt{ADJ}} only because ``red'' is repeatedly perturbed in one $x$), we prefer (2) templates that perturb various unique $x$.
We also prefer (3) finer-grained templates, to avoid being unnecessarily abstract (\eg to avoid abstracting ``not'' when it is the only \texttt{PART} changed.)

%$\hat{\xset}_i$

With these intuitions, we form the template selection as a weighted set coverage problem.
We see the union of counterfactuals for each $x$, $\hat{\xset}$, as the entire set of elements.
Then, each template $t \in T = {t_1,...,t_m}$ represents a subset of $\hat{\xset}$ that contains a number of counterfactuals $|t|$.
We define the weight as $w(t) = g(t) / |t|_x$, where $|t|_x$ quantifies the unique original $x$ covered by $t$, and $g(t)$ represents the sparsity of $t$ (heuristically decreasing from \texttt{text} to \texttt{POS}).
This way, templates that are too abstract or too focused on a certain $x$ are penalized by having a high weight. 
We use a classic greedy algorithm~\cite{vazirani2013approximation} to select a subset of $T^* \subset T$, such that the aggregated coverage is maximized, and the weight is minimized.

%\balance
%\input{sections/app_examples}

\end{document}